\documentclass[pdflatex,sn-basic,Numbered]{sn-jnl}

\usepackage{graphicx}%
\usepackage{multirow}%
\usepackage{amsmath,amssymb,amsfonts}%
\usepackage{amsthm}%
\usepackage{mathrsfs}%
\usepackage[title]{appendix}%
\usepackage{xcolor}%
\usepackage{textcomp}%
\usepackage{manyfoot}%
\usepackage{booktabs}%
\usepackage{algorithm}%
\usepackage{algorithmicx}%
\usepackage{algpseudocode}%
\usepackage{listings}%

\usepackage{xspace}
\usepackage{bm}
\usepackage{tabularx}
\usepackage{adjustbox}
\usepackage{rotating}
\usepackage{caption}

\newcommand{\exgnn}{\textsc{L2xGnn}\@\xspace}
\newcommand{\imle}{\textsc{I-MLE}\@\xspace}

\newcommand{\bA}{\bm{A}}
\newcommand{\bZ}{\bm{Z}}
\newcommand{\bX}{\bm{X}}

\newcommand{\btheta}{\bm{\theta}}
\newcommand{\bepsilon}{\bm{\epsilon}}

\newcommand{\bv}{\bm{v}}
\newcommand{\bu}{\bm{u}}
\newcommand{\by}{\bm{y}}

\newcommand{\optalgo}{\mathtt{opt}}

\algnewcommand{\Inputs}[1]{%
  \State \textbf{Input:}
  \Statex \hspace*{\algorithmicindent}\parbox[t]{.9\linewidth}{\raggedright #1}
}
\algnewcommand{\Initialize}[1]{%
  \State \textbf{Initialize:}
  \Statex \hspace*{\algorithmicindent}\parbox[t]{.9\linewidth}{\raggedright #1}
}

\DeclareMathOperator*{\argmax}{arg\,max}

\newcolumntype{C}{>{\centering\arraybackslash}m{4.8em}}
\newcolumntype{X}{>{\centering\arraybackslash}m{6.5em}}

\begin{document}

\title[L2XGNN: Learning to Explain Graph Neural Networks]{\exgnn: Learning to Explain Graph Neural Networks}


\author*[1]{\fnm{Giuseppe} \sur{Serra}}\email{serra@med.uni-frankfurt.de}
\equalcont{Work done at the University of Birmingham.}

\author[2]{\fnm{Mathias} \sur{Niepert}}\email{mniepert@gmail.com}

\affil*[1]{\orgname{Goethe University Frankfurt}, \orgaddress{\city{Frankfurt}, \country{Germany}}}

\affil[2]{\orgname{University of Stuttgart}, \orgaddress{\city{Stuttgart}, \country{Germany}}}


\abstract{
    Graph Neural Networks (GNNs) are a popular class of machine learning models. Inspired by the learning to explain (L2X) paradigm, we propose \exgnn, a framework for explainable GNNs which provides \textit{faithful} explanations by design. 
    \exgnn learns a mechanism for selecting explanatory subgraphs (motifs) which are exclusively used in the GNNs message-passing operations.  \exgnn is able to select, for each input graph, a subgraph with specific properties such as being sparse and connected. Imposing such constraints on the motifs often leads to more interpretable and effective explanations. 
    Experiments on several datasets suggest that \exgnn achieves the same classification accuracy as baseline methods using the entire input graph while ensuring that only the provided explanations are used to make predictions. Moreover, we show that \exgnn is able to identify motifs responsible for the graph's properties it is intended to predict.
}

\keywords{Graph-based machine learning, Interpretability, Explainability}



\maketitle
\section{Introduction}\label{sec:intro}
Graph Neural Networks (GNNs) are a widely used class of machine learning models. Since graphs occur naturally in several domains such as chemistry, biology, and medicine, GNNs have experienced widespread adoption. 
Following a trend toward building more interpretable machine learning models, there have been numerous recent proposals to provide explanations for GNNs. 
Most of the existing approaches provide post-hoc explanations starting from an already trained GNN to identify edges and node attributes that explain the model's prediction. However, as highlighted in \citet{faber2021comparing}, there might be some discrepancy between the ground-truth explanations and those attributed to the trained GNNs. Indeed, post-hoc explanations are often not able to faithfully represent the mechanisms of the original model \citep{rudin2018stop}. Unfortunately, the very definition of what constitutes a faithful explanation is still open to debate and there exist several competing positions on the matter. Recent work has also shown that post-hoc attribution methods are often not better than random baselines on the standard evaluation metrics for explanation accuracy and faithfulness~\citep{agarwal2022probing}.
Much fewer approaches have considered the problem of GNN explainability from an intrinsic perspective. In contrast to post-hoc methods, approaches with \textit{built-in} interpretability provide explanations during training by introducing new mechanisms, e.g. prototypes \citep{zhang2021protgnn}, stochastic attention \citep{miao2022interpretable}, or graph kernels \citep{feng2022kergnns}. Nonetheless, the introduction of new mechanisms to compute graph representations differ from standard GNNs computations. Therefore, the reasoning process of the above interpretable networks differ from the original GNN architectures making them not faithful by design. Our intent, instead, is to generate explanations for standard GNNs by keeping the computations as faithful as possible compared to the original network.

A recently proposed alternative to post-hoc methods is the learning to explain (L2X) paradigm~\citep{chen2018learning}. The core difference to post-hoc methods is that the models are trained to, in the forward pass,  discretely select a small subset of the input features as well as the parameters of a downstream model that uses only the selected features to make a prediction. The selected features are, therefore, faithful by design as they are the only ones used by the downstream model. Since the subset of features is sampled discretely, L2X requires a method for computing gradients of an expectation over a discrete probability distribution. \citet{chen2018learning} proposed a gradient estimator based on a relaxation of the discrete samples and tailored to the $k$-subset distribution. However, since the original work only considers the case of selecting \textit{exactly-k} features, directly applying prior methods to the graph learning tasks is not possible and requires significant changes. Thus, since prior work's gradient estimators do not work with arbitrary optimization problems but are restricted to the k-subset distribution, using the L2X paradigm for graphs is highly non-trivial.

With this work, we bring the L2X paradigm to graph representation learning. The important ingredient is a recently proposed method for computing gradients of an expectation over a complex exponential family distribution~\citep{niepert21imle}. The method facilitates approximate gradient backpropagation for models combining continuously differentiable GNNs with a black-box solver of combinatorial problems defined on graphs.  Crucially, this allows us to learn to sample subgraphs with beneficial properties such as being connected and sparse. Contrary to prior work, this also creates a dependency between the random variables representing the presence of edges. The proposed framework \exgnn, therefore, learns to select explanatory subgraph motifs and uses \emph{these and only these motifs} for its message-passing operations. 
To the best of our knowledge, this is the first method for learning to explain on standard GNNs. 
The proposed framework is extensible as it can work with any optimization algorithm for graphs imposing properties on the sampled subgraphs. 

We compare two different sampling strategies for obtaining sparse subgraph explanations resulting from two optimization problems on graphs: (1) the maximum-weight $k$-edge subgraph and (2) the maximum-weight $k$-edge connected subgraph problem. In line with \citet{faber2021comparing}, we decided to focus on explaining edges since the latter provide a more fine-grained information compared to nodes. 
We show empirically that \exgnn, when combined with a base GNN, does not lose accuracy on several benchmark datasets. Moreover, we evaluate the explanations quantitatively and qualitatively. We also analyze the ability of \exgnn to help in detecting shortcut learning which can be used for debugging the GNN.
Given the characteristics of the proposed method, our work improves model interpretability and increases the clarity of known black-box models, as GNNs, while maintaining competitive predictive capabilities. 

\section{Background}
\label{sec:background}
Let $\mathcal{G}(V, E)$ be a graph with $n=|V|$ the number of nodes. Let $\mathbf{X} \in \mathbb{R}^{n \times d}$ be the feature matrix that associates each node of the graph with a $d$-dimensional feature vector and let $\mathbf{A} \in \mathbb{R}^{n \times n}$ be  the adjacency matrix. GNNs have three computations based on the message passing paradigm \citep{hamilton2017inductive} which is defined as
\begin{equation}
    \label{eq:mpnn}
    \mathbf{h}_i^{\ell} = \gamma \left(\mathbf{h}_i^{\ell-1}, 
    \square_{j \in \mathcal{N}(v_i)} \phi \left(\mathbf{h}_i^{\ell-1}, \mathbf{h}_j^{\ell-1}, r_{ij} \right)\right),
\end{equation}
where $\gamma$, $\square$, and $\phi$ represent update, aggregation and message function respectively. \\
\textbf{Propagation step.} The message-passing network computes a message $m_{ij}^{\ell} = \phi(\mathbf{h}_i^{\ell-1}, \mathbf{h}_j^{\ell-1}, r_{ij})$ between every pair of nodes $(v_i, v_j)$. The function takes in input $v_i$'s and $v_j$'s representations $\mathbf{h}_i^{\ell-1}$ and $\mathbf{h}_j^{\ell-1}$ at the previous layer $\ell - 1$, and the relation $r_{ij}$ between the two nodes. \\
\textbf{Aggregation step.} For each node in the graph, the network performs an aggregation computation over the messages from $v_i$'s neighborhood $\mathcal{N}(v_i)$ to calculate an aggregated message $M_i^\ell = \square(\{m_{ij}^\ell \mid v_j \in \mathcal{N}(v_i)\})$. The definition of the aggregation function differs between methods \citep{hamilton2017inductive,velivckovic2018graph,xu2018powerful,duval2021graphsvx}. \\
\textbf{Update step.} Finally, the model non-linearly transforms the aggregated message $M_i^\ell$ and $v_i$'s representation from previous layer $\mathbf{h}_i^{\ell-1}$ to obtain $v_i$'s representation at layer $\ell$ as $\mathbf{h}_i^{\ell} = \gamma(M_i^\ell,\mathbf{h}_i^{\ell-1})$.
The final embedding for node $v_i$ after $L$ layers is $\mathbf{z}_i = \mathbf{h}_i^L$ and is used for node classification tasks. For graph classification, an additional readout function aggregates the node representations to obtain a graph representation $\mathbf{h}_G$. This function can be any permutation invariant function or a graph-level pooling function \citep{ying2018hierarchical,zhang2018end,lee2019self}. For Graph Isomorphism Networks (GINs) \citep{xu2018powerful}, for instance, the message passing operation for node $v_i$ is
\begin{equation} 
    \label{eq:gin_mp}
    \mathbf{h}_i^{\ell} = \gamma^{\ell} \left( 
    \left(1 + \epsilon^{\ell} \right) \cdot \mathbf{h}_i^{\ell-1} + 
    \sum_{j \in \mathcal{N}(v_i)} \mathbf{h}_j^{\ell-1} \right),
\end{equation}
where $\gamma$ represents a multi-layer perceptron (MLP), and $\epsilon$ denotes a learnable parameter. We will write $\mathbf{H}_\ell = \textsc{Gnn}_\ell (\bA, \mathbf{H}_{\ell-1})$ as a shorthand for the application of the $\ell^{\text{th}}$ layer of the GNN under consideration.

\section{Related Work}
There are several methods to explain the behavior of GNNs. Following \citet{yuan2020explainability}, explanatory methods for GNNs can be divided into several categories.

\noindent
\textbf{Gradient-based methods.} \citep{pope2019gradcam, baldassarre2019explainability, sanchez2020evaluating}. The main idea is to compute the gradients of the target prediction with respect to the corresponding input data. The larger the gradient values, the higher the importance of the input features. \\
\textbf{Perturbation-based methods.} \citep{ying2019gnnexplainer, luo2020pgexplainer, schlichtkrull2021interpreting, yuan2021subgraphx, perotti2022graphshap}. Here the objective is to study the models' output behavior under input perturbations. When the input is perturbed and we obtain an output comparable to the original one, we can conclude that the perturbed input information is not important for the current input. Inspired by causal inference methods, \citep{lin2021generative,lucic2022cf,tan2022learning} attempt to provide explanations based on factual and counterfactual reasoning. \\
\textbf{Surrogate methods.} \citep{huang2020graphlime, vu2020pgm, duval2021graphsvx, gui2022flowx}. First, these approaches generate a local dataset comprised of data points in the neighboring area of the input. The local dataset is assumed to be less complex and, consequently, can be analyzed through a simpler model. Then, a simple and interpretable surrogate model is used to capture local relationships that are used as explanations for the predictions of the original model. \\
\textbf{Decomposition methods.} \citep{schwarzenberg2019layerwise, schnake2020gnnlrp, feng2022degree}. These methods use decomposition rules to decompose the model predictions leading back to the input space. The prediction is considered as the target score. Then, starting from the output layer, the target score is decomposed at each preceding layer according to the decided decomposition rules. In this way, the initial target score is distributed among the neurons at every layer. Finally, the decomposed terms obtained at the input layer are associated to the input features and used as importance scores of the corresponding nodes and edges.  \\
\textbf{Model-level methods.} \citep{yuan2020xgnn}. Different from the instance-level methods above, these methods provide a general and high-level understanding of the models. In the context of GNNs, they aim at studying the input patterns that would lead to a certain target prediction. The generated explanations are general and provide a global understanding of the trained GNNs. \\
\textbf{Prototype-based methods.} \citet{zhang2021protgnn} propose ProtGNN, a new explanatory method based on prototypes to provide \textit{built-in} explanations, overcoming the limitations of post-hoc techniques. The explanations are obtained following case-based reasoning, where new instances are compared with several learned \emph{prototypes}. \\
\textbf{Concept-based methods.} \citet{magister2021gcexplainer} propose CGExplainer, a post-hoc explanatory methods for human-in-the-loop concept discovery. This concept representation learning method extracts concept-based explanations that allow the end-user to analyze predictions with a global view.

Among the methods categorized above, a similar approach in intent is presented in \citet{schlichtkrull2021interpreting}. The authors propose a post-hoc technique that learns how to remove the unnecessary edges through layer-wise edge masking. There are two main differences compared to our work: 1) the edge masking is learned from an already trained model, while we learn the edges to remove during training; 2) the edges are treated as independent binary random variables. In our case, instead, the optimization algorithm allows us to model the dependencies between edge variables.

Additional works face the explainability problem from different perspectives as explanation supervision \citep{gao2021gnes}, neuron analysis \citep{xuanyuan2022global}, and motif-based generation \citep{yu2022motifexplainer}. For a comprehensive discussion on methods to explain GNNs, we refer the reader to the survey \citep{yuan2020explainability}. In the following subsections, we provide a more detailed comparison with inherent interpretable methods and graph structure learning approaches.

\subsection{Comparison with Non-post-hoc Methods}
Among the plethora of post-hoc methods for graphs, ProtGNN~\citep{zhang2021protgnn}, KerGNN~\citep{feng2022kergnns}, and GSAT \cite{miao2022interpretable} are noteworthy exceptions. The first approach proposes a framework to generate explanations by comparing input graphs with prototypes learned during training, The second one combines graph kernels with the message passing paradigm to learn hidden graph filters. The latter, instead, leverages stochastic attention to select task-relevant subgraphs for interpretation. Although they all provide built-in explanations, given the introduction of new mechanisms to compute graph representations that differ from standard GNNs computations, the aforementioned approaches are not faithful by design (i.e., they do not reflect the reasoning process of the original backbone architecture). In contrast to these methods, our approach relies solely on standard GNNs, making it suitable to explain them faithfully. Additionally, in terms of explanatory capability, the learned prototypes are not directly interpretable and need to be matched to the closest training subgraphs to be human-understandable. Graph filters, instead, do not necessarily match existing patterns in the instance-based case. In both cases, the output can only provide a general idea of the important structures used by the model for prediction but fail at revealing precisely the instance-level explanation for each input graph.

\subsection{Comparison with Graph Structure Learning Approaches}
Recently, there have been related methods for learning the structure of graph neural networks. Following the taxonomy proposed in \citet{zhu2021deep}, the structure learning methods most related to \exgnn fall into the \textit{postprocessing} category, and more specifically, under the \textit{discrete sampling} subcategory. All existing methods use variants of the Gumbel-softmax trick which is limited in modeling complex distributions. Moreover, only when the straight-through version of the Gumbel-softmax trick is used, one can obtain truly discrete and not merely relaxed adjacency matrices in the forward pass. In contrast, \exgnn always samples purely discrete adjacency matrices. It is, to the best of our knowledge, the only method that allows us to model complex dependencies between the edge variables through its ability to integrate a combinatorial optimization algorithm on graphs. Other strategies include sampling edges between each pair of nodes from a Bernoulli distribution~\citep{franceschi2019learning} or sampling subgraphs for subgraph aggregation methods in a data-driven manner~\citep{qian2022ordered}. All these methods, however, are not concerned with the problem of explaining the behavior of GNNs explicitly. 

\subsection{Limitations of Prior Work}
When explaining GNNs, we distinguish between how the dataset was constructed and how the GNN makes its predictions. We refer to a \textit{responsible} motif when a dataset is created such that the presence or absence of it determines the class label of the graphs. Hence, the responsible motif represents the underlying evidence (ground truth) allowing us to discriminate among the labels that we hope the explanatory method will find \cite{faber2021comparing}. Instead, when a motif is responsible for the \textit{prediction} of a certain class label, we refer to the edges present in the motif as the ones \textit{causing} the prediction (\textit{causing} motif). Existing XAI methods for GNNs have several limitations and can lead to inconsistencies. In fact, there could be a mismatch between the responsible motif (ground truth), the actual motif used by the pre-trained model for its prediction (causing motif), and the one identified by the explanatory model (\textit{explanatory} motif)~\citep{duval2021graphsvx,faber2021comparing}. In contrast, in our work, we know that the prediction of the class label is caused by the explaining motif, as its selection by the upstream model caused the downstream model to make said prediction. As anticipated, we focus on the problem of identifying a subset of the edges as an explanation of the model's message-passing behavior. Hence, an explanation is equivalent to identifying a mask for the adjacency matrix of the original graph. Intuitively, an explanation can be \emph{accurate} and/or \emph{faithful}. It is accurate if it succeeds in identifying the edges in the \emph{input graph} responsible for the graph's class label – i.e., if the explanatory motif matches the responsible motif. This property can, for example, be evaluated with synthetic data where the class label of a graph is determined by the presence or absence of a particular substructure. An explanation is faithful if the edges identified as the explanation \emph{cause} the prediction of the GNN on an input graph – i.e., if the explanatory motif matches the causing motif. Contrary to measuring accuracy, there is no consensus on evaluating faithfulness. 

Recent work has proposed to measure unfaithfulness as the difference between the predictions of (1) the GNN on a perturbed adjacency matrix and (2) the GNN on the same perturbed adjacency matrix with edges removed by the explanation mask~\citep{pope2019gradcam,agarwal2022probing,agarwal2022explainable}.
We believe that this definition is problematic as the perturbation is typically implemented using a swap operation which replaces two existing edges $(a, b)$ and $(c, d)$ with two \emph{new} edges $(a, c)$ and $(b, d)$. Hence, these new edges are present in the unmasked adjacency matrix but not present in the masked one. It is, however, natural that the same GNN would predict highly different label distributions on these two graphs. For instance, consider a chemical compound where we remove and add new bonds. The resulting compounds and their properties can be chemically very different. Hence, contrary to prior work, we define a subgraph to be a faithful explanation, if it is a significantly smaller subgraph of the input graph and we know that \emph{only its structure} is used in the message-passing operations of equation~(\ref{eq:mpnn}).

\begin{figure*}[t]
\centering
\includegraphics[width=0.9\textwidth]{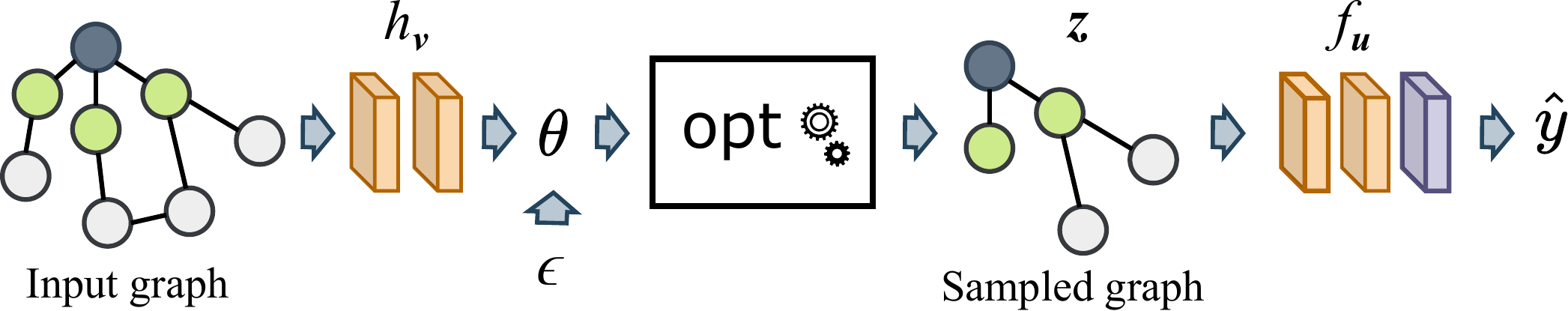} 
\caption{Workflow of the proposed approach. The upstream model $h_{\bm{v}}$ learns to assign weights $\theta_{\cdot,\cdot}$ for each edge in the input graph. The edge matrix $\btheta$ – perturbed with $\bepsilon$ – is then utilized as input by the optimization algorithm $\optalgo$ to sample a subgraph $\bm{z}$ with specific characteristics. Finally, the downstream model $f_{\bm{u}}$ uses \textit{only} the information about the sampled (sub)graph to make a prediction.}
\label{fig:imle_workflow}
\end{figure*}

\section{Learning to Explain Graph Neural Networks}
We propose a method that learns both (i) the parameters of a graph generative model and (ii) the parameters of a GNN operating on sparse subgraphs approximately sampled from said generative model in the forward pass. In line with prior work on learning to explain~\citep{chen2018learning}, the maximum probability subgraph is then used at test time to make the prediction and, therefore, serves as the faithful explanation. Since we aim to sample graphs with certain properties (e.g., connected subgraphs) we need a new approach to sampling and gradient estimation. 
Contrary to prior work on edge masking \citep{schlichtkrull2021interpreting} which treats edges as independent binary random variables, we use a recently introduced method for backpropagating through optimization algorithms. This allows us to select subgraphs with specific properties and, therefore, to explicitly model dependencies between edge variables.

Intuitively, our approach consists of three main components. In the first part, an upstream model $h_{\bm{v}}$ learns the edge weights $\theta_{(i,j)}$ for each edge $(i,j)$ belonging to the given input graph. In the subsequent component, the learned edge matrix $\btheta$ is given as input to an optimization algorithm $\optalgo$. The algorithm considers the weights $\btheta$ as unnormalized probabilities to sample discretely a new adjacency matrix $\bZ$. Finally, the resulting sampled subgraph $\bm{z}$ is used in the last component, the downstream model $f_{\bm{u}}$, to make the final prediction. A graphical representation of our approach is presented in Figure \ref{fig:imle_workflow}. Considering the proposed workflow, we can identify two main challenges related to our method: a) how to learn $\btheta$ such that we can improve the selection of the subgraph $\bm{z}$; b) how to estimate and backpropagate the gradient through a discrete component (i.e., $\optalgo$). In the following subsections, we will explain our framework in more detail and provide technical solutions for the introduced challenges. In Subsection \ref{sec:problem_statement}, we formalize the problem and describe rigorously our framework. In Subsection \ref{sec:imle}, we describe the gradient estimation method used in this work. Finally, in Subsection \ref{sec:l2xgnn}, we detail how to use and adapt the introduced concepts to work explaining GNNs.

\subsection{Problem Statement and Framework} \label{sec:problem_statement}

We aim to jointly learn the parameters of a probability distribution over subgraphs \emph{with certain properties} and the parameters of a GNN operating on graphs sampled from said distribution in the context of the graph classification problem. 
Here, the training data consists of a set of triples $\{(\mathbf{A}, \mathbf{X}, \mathbf{y})_j\}, j \in \{1, ..., N\}$, where $\mathbf{A}$ is an $n \times n$ binary adjacency matrix, $\mathbf{X} \in \mathbf{R}^{n \times d}$ a node attribute matrix with $d$ the number of node attributes, and $\mathbf{y}$ the target graph label. First, we have a learnable function $h_{\bm{v}}: \mathcal{A} \times \mathcal{X} \rightarrow \Theta$ where $\mathcal{A}$ is the set of all $n \times n$ adjacency matrices, $\mathcal{X}$ the set of all attribute matrices, $\bv$ are the parameters of $h$, and $\Theta$ the set of possible edge parameter values. The function, which we refer to as the upstream model, maps the adjacency and attribute matrix to a matrix of edge weights $\btheta \in \mathbb{R}^{n \times n}$. Intuitively, $\btheta_{i,j}$ is the prior probability of edge $(i, j)$.

Next, we assume an algorithm $\optalgo: \Theta \rightarrow \mathcal{A}$ which returns the (approximate) solutions to an optimization problem on edge-weighted graphs. Examples of such optimization problems are the maximum-weight spanning tree or the maximum-weight  $k$-edge connected subgraph problems. The optimization algorithm is treated as a black box. One can choose the optimization problem according to the application's requirements. We have found, for instance, that the connected subgraphs lead to better explanations in the domain of chemical compound classification. Contrary to prior work, the optimization problem creates a dependency between the binary variables modeling the edges. 

For every binary adjacency matrix $\bZ \in \mathcal{A}$, we write $\bZ \in \mathcal{F}$ if and only if the adjacency matrix is a feasible solution (not necessarily an optimal one) of the chosen optimization problem.
We can now define a discrete exponential family distribution as 
\begin{equation} \label{def-constrained-exp-family}
p(\bZ; \btheta) = \left\lbrace
\begin{array}{ll}
     \exp\left( \langle \bZ, \btheta \rangle_{F} - B(\btheta) \right) & \text{if } \bZ \in \mathcal{F}, \\
     0 & \text{otherwise.} 
\end{array} \right.
\end{equation}
where $\langle \cdot, \cdot \rangle_{F}$ is the Frobenius inner product and $B(\btheta)$ is the log-partition function defined as 
$$B(\btheta) = \log\left(\sum_{\bZ \in \mathcal{F}} \exp \left( \langle \bZ, \btheta \rangle_{F} \right)\right).$$
Hence, $p$ is a probability distribution over adjacency matrices that are feasible solutions to the optimization problem under consideration. Each feasible adjacency matrix's probability mass is proportional to the product of its edge weights. For example, if the optimization problem is the maximum-weight $k$-edge connected subgraph problem, the distribution assigns a non-zero probability mass to all adjacency matrices of graphs that have $k$ edges and are connected. 

Given an optimization problem, we would like to sample exactly from the above probability distribution $p(\bZ; \btheta)$. Unfortunately, this is intractable since computing the log-partition function is in general NP-hard. However, as in prior work~\citep{niepert21imle}, we can use perturb-and-MAP~\citep{Papandreou:2011} to \emph{approximately} sample from the above distribution as follows. 
Let $\bepsilon \sim \rho(\bepsilon)$ be a $n \times n$ matrix of appropriate random variables such as those following the Gumbel distribution. We can then \emph{approximately} sample an adjacency matrix $\bZ$ from $p(\bZ; \btheta)$  by computing
$$\bZ = \optalgo(\btheta + \bepsilon).$$
Hence, we can approximately sample by perturbing the edge weights (unnormalized probabilities) $\btheta$ and by applying the optimization algorithm to these perturbed weights. 

In the final part of the model (the downstream model), we use the sampled $\bZ$ as the input adjacency matrix to a message-passing neural network $f_{\bu}: \mathcal{A} \times \mathcal{X} \rightarrow \mathcal{Y}$ computing $\hat{\by} = f_{\bu}(\bZ, \bX)$.

In summary, we have the following model architecture for training input data $(\bA, \bX, \by)$:
\begin{align}
\label{eq:model_input}
    \btheta & = h_{\bm{v}}(\bA, \bX)  & \mbox{ with } \ \bA \in \mathcal{A}, \bm{X} \in \mathcal{X}, \\
\label{eq:model_opt}
    \bZ & = \optalgo(\btheta + \bepsilon)  & \mbox{ with }  \ \bepsilon \sim \rho(\epsilon), \bepsilon \in \mathbb{R}^{n \times n}, \\
\label{eq:model_output}
    \hat{\by} & = f_{\bm{u}}(\bZ, \bX)  & \mbox{ with } \  \hat{\by} \in \mathcal{Y}, f_{\bm{u}}: \mathcal{A} \times \mathcal{X} \rightarrow \mathcal{Y}. 
\end{align} 
Figure~\ref{fig:imle_workflow} illustrates the architecture.
With $\bm{\omega} = (\bm{u},\bm{v})$ the learnable parameters of the model and the target variable $\by$ the loss is now defined as:
\begin{equation}
\label{eq:loss}
    L(\bA, \bX, \by;\bm{\omega}) = \mathbb{E}_{\bepsilon \sim \rho(\epsilon)}[\ell(f_{\bm{u}}(\bZ, \bX),\by)],
\end{equation}
with $\bZ = \optalgo(\btheta + \bepsilon) $, $\btheta=h_{\bm{v}}(\bA,\bX)$, and  $\ell: \mathcal{Y} \times \mathcal{Y} \rightarrow \mathbb{R}^{+}$  a point-wise loss function. 
The gradient of $L$ wrt $\bm{u}$ is
\begin{equation*}
\label{eq:gradient_u}
    \nabla_{\bm{u}} L(\bA, \bX, \by;\bm{\omega}) =
    \mathbb{E}_{} [\partial_{\bm{u}}f_{\bm{u}}(\bZ, \bX)^{\intercal}
    \nabla_{\by} \ell (\hat{\by},\by)]
\end{equation*}
which can be estimated by Monte-Carlo sampling. In contrast, the gradient of $L$ with respect to $\bm{v}$ is:
\begin{equation*}
\label{eq:gradient_v}
    \nabla_{\bm{v}} L(\bA, \bX, \by;\bm{\omega})  =
    \partial_{\bm{v}}h_{\bm{v}}(\bA, \bX)^{\intercal}
    \nabla_{\btheta} L(\bA, \bX, \by;\bm{\omega}),
\end{equation*}
where the  challenge is to estimate  $\nabla_{\btheta} L(\bA, \bX, \by;\bm{\omega}) = \nabla_{\btheta}\mathbb{E}_{\bepsilon \sim \rho(\epsilon)}[\ell(f_{\bm{u}}(\bZ, \bX),\by)]$ because $\bZ = \optalgo(\btheta + \bepsilon)$ is not continuously differentiable wrt $\btheta$. While it would be possible to use the score function estimator, its high variance makes it less competitive in practice~\citep{niepert21imle}.

\subsection{Implicit Maximum-Likelihood Learning} \label{sec:imle}

The variant of \imle we use in this work estimates $\nabla_{\btheta} L(\bA, \bX, \by;\bm{\omega})$ by implicitly creating a target distribution $q(\bZ; \btheta')$ using perturbation-based implicit differentiation~\citep{domke2010implicit}. Here, the parameters $\btheta$ are moved in the direction of $-\nabla_{\bZ} \ell (f_{\bu}(\bA, \bX),\by))$, the negative gradient of the downstream loss with respect to the sampled adjacency matrix $\bZ$, to construct $\btheta'$
\begin{equation}
\label{eq:target_distribution}
    q(\bZ;\bm{\theta}^{\prime}) :=
    p(\bZ;\bm{\theta} - \lambda \nabla_{\bZ} \ell (f_{\bu}(\bZ, \bX),\by))
\end{equation}
with $\bZ = \optalgo(\btheta + \bepsilon)$ and $\lambda > 0$ the strength of the perturbation. Intuitively, by moving the weights $\btheta$ into the direction of the negative gradients of $\bZ$, the resulting distribution $q$ is more likely to generate samples with a lower downstream loss.
We approximate $\nabla_{\btheta} L(\bA, \bX, \by;\bm{\omega})$ with Monte Carlo estimates of the gradients of the KL divergence between $p$ and $q$:
\begin{equation}
\label{eq:approx_gradient}
    \nabla_{\btheta} L(\bA, \bX, \by;\bm{\omega}) \approx 
    \frac{1}{\lambda} \left(\optalgo(\btheta + \bepsilon) - 
    \optalgo(\btheta^{\prime} + \bepsilon)\right).
\end{equation}
In other words, $\nabla_{\btheta} L(\bA, \bX, \by;\bm{\omega})$ is approximated by the difference between an approximate sample from $p(\bZ;\btheta)$ and an approximate sample from $q(\bZ;\btheta^{\prime})$. In this way we move the distribution $p(\bZ;\btheta)$ closer to $q(\bZ;\btheta^{\prime})$. 

\subsection{L2XGNN: Learning to Explain GNNs with I-MLE} \label{sec:l2xgnn}

We now describe the class of \exgnn models we use in the experiments. First, we need to define the function $h_{\bm{v}}(\bA, \bX)$. Here we use a standard GNN (see Equation~\ref{eq:mpnn}) to compute for every node $i$ and every layer $\ell$ the vector representation $\mathbf{h}_i^{\ell} = h_{\bm{v}}(\bA, \bX)_{i,1:d}$.
We then compute the matrix of edge weights by taking the inner product between each pair of node embeddings. More formally, we compute $\btheta_{i,j} = \langle \mathbf{h}^{\ell}_i, \mathbf{h}_j^{\ell} \rangle$ for some fixed $\ell$. Typically, we choose $\ell=1$. 

In this work, we sample the noise perturbations $\bm{\epsilon}$  from the sum of Gamma distribution~\citep{niepert21imle}.
Other noise distributions such as the Gumbel distribution are possible. 

\subsubsection{Sampling Constrained Subgraphs}
An advantage of the proposed method is its ability to integrate any graph optimization problem as long as there exists an algorithm $\optalgo$ for computing (approximate) solutions. In this work, we focus on two optimization problems: (1) The maximum-weight $k$-edge subgraph and (2) the maximum-weight $k$-edge connected subgraph problems. The former aims to find a maximum-weight subgraph with $k$ edges. The latter aims to find a \emph{connected} maximum-weight subgraph with $k$ edges. Other optimization problems are possible but we found that sparse and connected subgraphs provide a good efficiency-effectiveness trade-off.  

Computing maximum weight $k$-edge subgraphs is highly efficient as we only need to select the $k$ edges with the maximum weights. 
In order to compute \emph{connected} $k$-edge subgraphs we use a greedy approach. First, given a number $k$ of edges, we select a single edge $e_{i,j}$ with the highest weight $\btheta_{i,j}$ from the input graph. At every iteration of the algorithm, we select the next edge such that it (a) is connected to a previously selected edge and (b) has the maximum weight among all those connected edges. A more detailed description of the greedy algorithm is given in Algorithm \ref{alg:cap}. 

\begin{algorithm}[t]
\caption{Greedy algorithm $\optalgo$ for the maximum-weight $k$-edge connected subgraph problem.}
\label{alg:cap}
\begin{algorithmic}
\Inputs{Input graph $\mathcal{G} = (V, E)$ \\
        Number of edges $k$ \\ Edge weights $\btheta$ \\
        }
\Initialize{
            $e = \argmax_{e_{i,j}} \btheta_{i,j}$ \\
            Set of selected edges $S = \{ e \}$ \\
            Set of edges adjacent to selected edge $N = \mathcal{N}(e)$
            }
\While{$|S| < k$ and $|N| > 0$}
    \State $e = \argmax_{e_{i,j} \in N} \btheta_{i,j}$ 
    \State $S = S \cup \{e\}$ 
    \State $N = N \cup \mathcal{N}(e)$
    \State $N = N - S$
\EndWhile
\State \textbf{Return:} Adjacency matrix $\bZ$ of the subgraph induced by the set of selected edges $S$.
\end{algorithmic}
\end{algorithm}

Finally, we need to define the function $f_{\bu}$ (the downstream function) of the proposed framework. Here, we again use a message-passing GNN that follows the update rule
\begin{equation}
    \label{eq:mpnn-2}
    \mathbf{h}_i^{\ell} = \gamma \left(\mathbf{h}_i^{\ell-1}, 
    \square_{j \in \mathcal{N}(v_i)} \phi \left(\mathbf{h}_i^{\ell-1}, \mathbf{h}_j^{\ell-1}, r_{ij} \right)\right).
\end{equation}
The neighborhood structure $\mathcal{N}(\cdot)$, however, is defined through the output adjacency matrix $\bZ$ of the optimization algorithm $\optalgo$
\begin{equation}
    j \in \mathcal{N}(v_i) \Longleftrightarrow \bZ_{i, j} = \bZ_{j, i} = 1.
\end{equation}

Hence, if after the subgraph sampling, there exists a node $v_i$ which is an isolated node in the adjacency matrix $\bZ$, that is, $\bZ_{i,j} = \bZ_{j,i} = 0 \ \forall j \in \{1, ..., n\}$, the embedding of the node will not be updated based on message passing steps with neighboring nodes. This means that, for \textit{isolated} nodes, the only information used in the downstream model is the one from the nodes themselves. Conceptually, $\bZ$ works as a mask over the messages $m_{ij}^\ell$ computed at each layer $\ell$.

The adjacency matrix $\bZ$ is then used in all subsequent layers of the GNN. In particular, for one layer $\ell$ we have
\begin{equation}
    \label{eq:gnn_layer}
    \mathbf{H}_\ell = \textsc{Gnn}_\ell (\bA \odot \bZ, \mathbf{H}_{\ell-1}),
\end{equation}
where $\odot$ is the Hadamard product. 
Finally, the remaining part of the \exgnn network for the graph classification is 
\begin{equation}
    \label{eq:classifier}
    \mathbf{h}_{G} = \text{Pool}(\mathbf{H}_\ell) \qquad \hat{\bm{y}} = \eta(\mathbf{h}_G),
\end{equation}
where we use a global pooling operator to generate the (sub)graph representation $\mathbf{h}_G$ that will then be used by the MLP network $\eta(\cdot)$ to output a probability distribution $\hat{\by}$ over the class labels. Finally, a loss function is applied whose gradients are used to perform backpropagation. At test time, we use the maximum-probability subgraph for the explanation and prediction, that is, we do not perturb at test time.

\section{Experiments} 

First, we evaluate the predictive performance of the model compared to baselines. Second, we qualitatively and quantitatively analyze the explanatory subgraphs for datasets for which we know the ground-truth motifs. Finally, we analyze whether the generated output can be helpful for model debugging purposes. We report several ablation studies to investigate the effects of different model choices on the results in the supplementary material. For the remainder of the manuscript, we use \exgnn$_\texttt{dsc}$ and \exgnn for referring to the maximum-weight k-edge subgraph and to the maximum-weight k-edge \textit{connected} subgraph problem respectively. The code for reproducing our experiments is available \href{https://github.com/GiuseppeSerra93/L2XGNN}{here}.

\subsection{Datasets and Settings} 

\subsubsection{Datasets} To understand the change in the predictive capabilities of the base models when integrating \exgnn, we use six real-world datasets from different domains (biology, social networks) for graph classification tasks: MUTAG \citep{debnath1991structure}, PROTEINS \citep{borgwardt2005protein}, YEAST \citep{yan2008mining}, IMDB-BINARY, IMDB-MULTI \citep{yanardag2015deep}, and DD \citep{rossi2015network}. In Table~\ref{tab:statistics}, we report the statistics of the datasets used for graph classification tasks. For a comprehensive evaluation, we include datasets with different characteristics, such as a larger number of graphs or a larger number of nodes and edges.

\begin{table}[h]
  \caption{Statistics of the datasets.}
  \label{tab:statistics}

\begin{tabular}{lcccc}
\toprule
\multicolumn{1}{c}{Number of}   & Nodes (avg)  &  Edges (avg) &   Graphs   & Classes \\
\midrule
DD       & 284.32   & 715.66    & 1178      & 2     \\
MUTAG    & 17.93    & 19.79     & 188       & 2     \\
IMDB-B   & 19.77    & 96.53     & 1000      & 2     \\
IMDB-M   & 13.00    & 65.94     & 1500      & 3     \\
PROTEINS & 39.06    & 72.82     & 1113      & 2     \\
YEAST    & 21.54    & 22.84     & 79601     & 2     \\
      
\bottomrule
\end{tabular}
\end{table}

\begin{wrapfigure}{R}{0.16\textwidth}
\vspace{-20pt}
\includegraphics[width=0.16\textwidth]{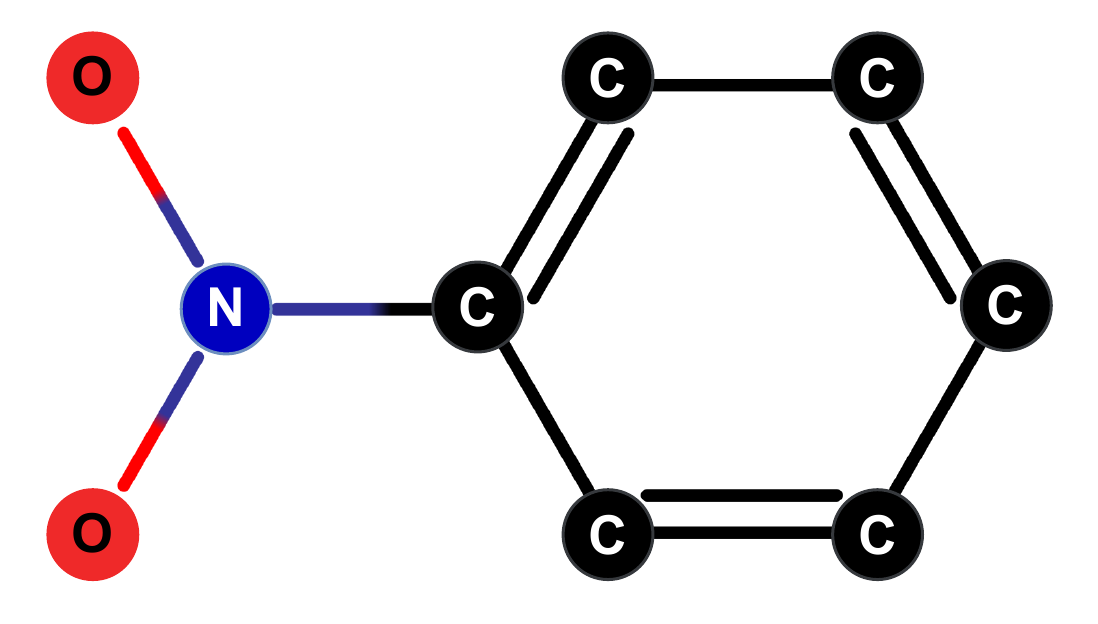} 
\caption{Benzene-NO$_2$ motif.}
\vspace{-30pt}
\label{fig:ground_truth}
\end{wrapfigure}

To quantitatively evaluate the quality of the explanations, we use datasets that include ground-truth edge masks. In particular, we use MUTAG$_0$ and BA2Motifs. MUTAG$_0$ is a dataset introduced in \citet{tan2022learning} which contains the benzene-NO$_2$ (i.e., a carbon ring with a nitro group (NO$_2$) attached) as the only discriminative motif between positive and negative labels. A graphical representation of the benzene-NO$_2$ compound is given in Figure \ref{fig:ground_truth}. BA2Motifs is a synthetic dataset that was first introduced in \citet{luo2020pgexplainer}. The base graphs are Barabasi-Albert (BA) graphs. 50\% of the graphs are augmented with a \textit{house-motif} graphs, the rest with a \textit{5-node cycle motif}.  The discriminative subgraph leading to different predictions is the motif attached to the BA graph.

\subsubsection{Experimental Settings} To evaluate the quality of our approach, we use \exgnn with several GNN base models including GCN \citep{kipf2016semi}, GIN \citep{xu2018powerful} and GraphSAGE \citep{hamilton2017inductive}. We compare the results when using the original model and when the same model is combined with our XAI method. 
For model selection and evaluation, to fairly compare the methods, we follow a previously proposed  protocol\footnote{\url{https://github.com/pyg-team/pytorch_geometric/tree/master/benchmark/kernel}}. We perform a 10-fold cross validation where the hyperparameter selection is done according to the validation accuracy. The selection is performed for the number of layers (L) $[1,2,3,4]$ and the number of hidden units (H) $[16, 32, 64, 128]$. For both parameters, the selected numbers represent a standard range of values to decide the characteristics of the backbone architecture. For a fair comparison with the backbone architectures, we select the best configuration for each dataset, and we integrate our approach into the best model. Instead of fixing a value $k$ for each input graph, we compute $k$ based on a ratio $R$ of edges to be kept. Once the hyperparameters of the default model are found, we select the best ratio $R$ (in terms of percentage of edges to keep) from the set of values $[0.4, 0.5, 0.6, 0.7]$ based again on the validation accuracy. We do not include extreme values for two reasons: (1) smaller values for $R$ lead to reduced predictive capabilities and not meaningful explanatory subgraphs; and (2) higher values would not remove enough edges compared to the original input. Finally, we choose the perturbation intensity $\lambda$ from the values $[10, 100, 1000]$ taken from the original paper \cite{niepert21imle}.

Experiments were run on a single Linux machine with Intel Core i7-11370H @ 3.30GHz, 1 GeForce RTX 3060, and 16 GB RAM. The best hyperparameter configuration for each model and dataset used for graph classification tasks is reported in Table~\ref{tab:hyperparam}. First, for the backbone architectures, we consider the number of layers $[1,2,3,4]$ and the number of hidden units $[16,32,64,128]$. Then, for \exgnn, we select the ratio $R$ from  $[0.4,0.5,0.6,0.7]$ and the perturbation intensity $\lambda$ from $[10,100,1000]$.

\begin{table}[h]
  \caption{Hyperparameter settings for graph classification tasks. H and L represent the number of hidden units and the number of layers respectively.}
  \label{tab:hyperparam}
\begin{tabular}{lcccccccccccc}
\toprule
\multicolumn{1}{l}{\multirow{2}{*}{Dataset}} & \multicolumn{4}{c}{GCN} & \multicolumn{4}{c}{GIN} & \multicolumn{4}{c}{GraphSAGE} \\ 
\cmidrule(r){2-5}\cmidrule(r){6-9}\cmidrule(r){10-13}

\multicolumn{1}{c}{}   & H    &  L   & $R$  & $\lambda$ &
                         H    &  L   & $R$  & $\lambda$ &
                         H    &  L   & $R$  & $\lambda$\\
\cmidrule(r){2-5}\cmidrule(r){6-9}\cmidrule(r){10-13}
DD    & 128 & 2 & 0.6 & 10 & 64 & 1 & 0.5 & 100 & 128 & 1 & 0.6 & 100\\
MUTAG & 128 & 3 & 0.6 & 1000 & 128 & 4 & 0.5 & 10 & 128 & 3 & 0.4 & 10\\
IMDB-B  & 128 & 3 & 0.4 & 100 & 64 & 3 & 0.4 & 1000 & 64 & 1 & 0.4 & 10\\
IMDB-M  & 64 & 3 & 0.6 & 10 & 128 & 4 & 0.6 & 10 & 64 & 1 & 0.4 & 100\\
PROTEINS  & 128 & 3 & 0.7 & 100 & 128 & 4 & 0.5 & 10 & 64 & 3 & 0.4 & 10\\
YEAST    & 128 & 3 & 0.6 & 10 & 128 & 3 & 0.6 & 10 & 32 & 4 & 0.5 & 100\\
\bottomrule
\end{tabular}
\end{table}

\vspace{-10pt}
\subsection{Empirical Results}
\subsubsection{Graph Classification Comparison with Base GNNs} \label{sec:graph_classification}
Following the experimental procedure proposed in \citet{zhang2021protgnn}, Table \ref{tab:accuracy} lists the results of using \exgnn with base GNN architectures for graph classification tasks. We observe that \exgnn is competitive and often even outperforms the base GNN models on the benchmark datasets. The primary goal of this work is not to provide a better predictive model, but to provide faithful explanation masks while maintaining similar predictive performance. To prove this point, we perform a paired t-test via 5x2 cross-validation with significant level $\alpha=0.05$  \citep{dietterich1998approximate} (see Appendix \ref{app:5x2} for more details). The test indicates there is \textit{no statistically significant difference} between the base models and their explainable counterpart (either in the connected or disconnected version). This analysis is important since inherent interpretable networks are known for creating a trade-off with the predictive capabilities of the model, and practitioners may not be willing to sacrifice the prediction accuracy for increased transparency \citep{miao2022interpretable}. 

\begin{table}[t]
\caption{Prediction test accuracy (\%) for graph classification tasks over ten runs.} \label{tab:accuracy}

\begin{tabular}{lccccccccc}
\toprule
\multicolumn{1}{l}{\multirow{2}{*}{Method}} &
\multicolumn{6}{c}{Dataset} \\

\cmidrule(r){2-7}
\multicolumn{1}{c}{} & \multicolumn{1}{c}{DD} & \multicolumn{1}{c}{MUTAG} & \multicolumn{1}{c}{IMDB-B} & \multicolumn{1}{c}{IMDB-M} & \multicolumn{1}{c}{PROTEINS} & \multicolumn{1}{c}{YEAST} \\

\cmidrule(r){2-2}\cmidrule(r){3-3}\cmidrule(r){4-4}\cmidrule(r){5-5}\cmidrule(r){6-6}\cmidrule(r){7-7}
GCN & \textbf{72.0} $\pm$ 2.4 & 73.4 $\pm$ 8.3 & 73.1 $\pm$ 3.2 & 50.0 $\pm$ 2.8 & 71.8 $\pm$ 4.4 & 88.1 $\pm$ 0.1 \\

$\textsc{L2xGcn}_{dsc}$ & 71.9 $\pm$ 3.1 & 73.9 $\pm$ 11.1 & 66.0 $\pm$ 5.4 & \textbf{50.3} $\pm$ 3.2 & 71.1 $\pm$ 3.4 & \textbf{88.2} $\pm$ 0.2 \\

\textsc{L2xGcn} & 71.9 $\pm$ 3.6 & \textbf{74.5} $\pm$ 8.2 & \textbf{73.4} $\pm$ 4.7 & 49.0 $\pm$ 2.2 & \textbf{72.0} $\pm$ 5.3 & 88.1 $\pm$ 0.1 \\

\cmidrule(r){2-2}\cmidrule(r){3-3}\cmidrule(r){4-4}\cmidrule(r){5-5}\cmidrule(r){6-6}\cmidrule(r){7-7}
GIN & 72.2 $\pm$ 2.7 & \textbf{82.7} $\pm$ 5.1 & 72.1 $\pm$ 5.0 & \textbf{49.0} $\pm$ 4.7 & 70.8 $\pm$ 4.5 & \textbf{88.3} $\pm$ 0.1 \\

$\textsc{L2xGin}_{dsc}$ & \textbf{73.9} $\pm$ 5.1 & 81.4 $\pm$ 9.2 & 65.0 $\pm$ 5.0 & 48.8 $\pm$ 3.2 & 68.5 $\pm$ 2.9 & 88.2 $\pm$ 0.1 \\

\textsc{L2xGin} & 72.0 $\pm$ 3.0 & 82.5 $\pm$ 7.8 & \textbf{72.4} $\pm$ 4.5 & 47.9 $\pm$ 3.5 & \textbf{70.9} $\pm$ 3.4 & 88.0 $\pm$ 0.2 \\

\cmidrule(r){2-2}\cmidrule(r){3-3}\cmidrule(r){4-4}\cmidrule(r){5-5}\cmidrule(r){6-6}\cmidrule(r){7-7}
Graph\textsc{Sage} & 72.1 $\pm$ 3.9 & 73.4 $\pm$ 7.5 & 72.2 $\pm$ 4.8 & 50.7 $\pm$ 3.7 & 71.3 $\pm$ 5.1 & \textbf{88.2} $\pm$ 0.1 \\

$\textsc{L2xGsg}_{dsc}$ & \textbf{72.7} $\pm$ 3.8 & 75.1 $\pm$ 7.7 & \textbf{73.8} $\pm$ 2.8 & 50.6 $\pm$ 3.2 & \textbf{71.3} $\pm$ 4.1 & 88.0 $\pm$ 0.1 \\

\textsc{L2xGsg} & 72.5 $\pm$ 3.9 & \textbf{79.8} $\pm$ 8.1 & 73.0 $\pm$ 4.1 & \textbf{50.8} $\pm$ 2.7 & 70.7 $\pm$ 4.6 & 88.1 $\pm$ 0.2 \\
\bottomrule

\end{tabular}
\end{table}

\subsubsection{Explanation Accuracy}
We compare the proposed method with popular post-hoc explanation techniques including GNN-Explainer \citep{ying2019gnnexplainer}, PGE-Explainer \citep{luo2020pgexplainer}, GradCAM \citep{pope2019gradcam}, GNN-LRP \citep{schnake2020gnnlrp}, and SubgraphX \citep{yuan2021subgraphx}\footnote{Implementations taken from the \textsc{Dig} library \citep{liu2021dig}.}. We train a 3-layer GIN for $200$ epochs with hidden dimensions equal to $64$ and a learning rate equal to $0.001$. We save the best model according to the validation accuracy and we compare it with the post-hoc techniques. In our case, we integrate \exgnn into the same architecture and learn the edge masking during training as described before. We report the graph classification results for the two datasets in the appendix. In Table \ref{tab:expl_accuracy}, we report the explanation accuracy evaluation with respect to the ground-truth motifs in comparison with post-hoc techniques for 5 different data splits. The explanation problem is formalized as a binary classification problem, where the edges belonging to the ground-truth motif are treated as positive labels. We observe that \exgnn obtains better or the same results as the considered explanatory models. While for the post-hoc explanation techniques we cannot guarantee that the GNNs use exclusively the explanation subgraphs for the prediction \citep{yuan2020explainability}, our method, by providing \textit{faithful} explanations, overcomes this limitation. It is exactly the provided explanation that is used in the message-passing operations of \exgnn.

\begin{table}[t]
\centering
\caption{Evaluation of explanation accuracy (\%) on synthetic graph classification datasets using a 3-layer GIN architecture. The lowest standard deviation for each metric is underlined. With the exception of \exgnn, none of the approaches can guarantee \textit{faithful} explanations where the explanation is exclusively used during message passing operations. }
\label{tab:expl_accuracy}
\begin{tabular}{lcccc}
\toprule
\multicolumn{1}{l}{Dataset} & \multicolumn{4}{c}{BA-2MOTIFS}\\ 
\cmidrule(r){2-5}
\multicolumn{1}{c}{}   & Acc.    & Pr.   & Rec. & F$_1$  \\
\midrule
GNN-Exp. & 44.6  \footnotesize$\pm$ 2.4 & 22.7  \footnotesize$\pm$ 0.9 & 62.9  \footnotesize$\pm$ 3.3 & 32.9  \footnotesize$\pm$ \underline{1.0} \\
GradCAM & 77.1  \footnotesize$\pm$ 11.5 & 50.1  \footnotesize$\pm$ 16.1 & 72.4  \footnotesize$\pm$ 23.2 & 59.0  \footnotesize$\pm$ 19.0 \\
PGE-Exp. & 36.7  \footnotesize$\pm$ 18.9 & 17.5  \footnotesize$\pm$ 5.9 & 66.6  \footnotesize$\pm$ 22.5 & 27.7  \footnotesize$\pm$ 9.4\\
GNN-LRP & 77.3  \footnotesize$\pm$ 2.5 & 34.3 \footnotesize$\pm$ 16.8 & 36.4  \footnotesize$\pm$ 19.7 & 33.0  \footnotesize$\pm$ 15.7 \\
\textit{SubgraphX} & \textbf{81.5}  \footnotesize$\pm$ 5.6 & \textbf{54.0} \footnotesize$\pm$ 12.9 & 74.2  \footnotesize$\pm$ 16.5 & 60.4  \footnotesize$\pm$ 14.2 \\

\midrule
\textsc{L2xGin} & 78.0  \footnotesize$\pm$ \underline{0.6} & 49.5  \footnotesize$\pm$ \underline{0.8} & 90.2  \footnotesize$\pm$ \underline{1.4} & 63.8  \footnotesize$\pm$ \underline{1.0} \\
$\textsc{L2xGin}_{dsc}$  & 80.0  \footnotesize$\pm$ 1.2 & 52.1  \footnotesize$\pm$ 1.6 & \textbf{94.7}  \footnotesize$\pm$ 2.7 & \textbf{67.1}  \footnotesize$\pm$ 2.0 \\

\midrule

\multicolumn{1}{l}{Dataset} & \multicolumn{4}{c}{MUTAG$_{0}$}\\ 
\cmidrule(r){2-5}
\multicolumn{1}{c}{}   & Acc.    & Pr.   & Rec. & F$_1$ \\
\midrule
GNN-Exp. & 47.4 \footnotesize$\pm$ 2.3 & 42.2  \footnotesize$\pm$ \underline{2.4} & 69.2 \footnotesize$\pm$ \underline{2.4} & 50.2 \footnotesize$\pm$ \underline{2.1} \\
GradCAM & \textbf{78.0} \footnotesize$\pm$ \underline{1.3} & \textbf{85.6}  \footnotesize$\pm$ 2.8 & 60.8 \footnotesize$\pm$ 3.9 & 68.8 \footnotesize$\pm$ 2.2 \\
PGE-Exp. & 65.0 \footnotesize$\pm$ 9.6 & 57.3  \footnotesize$\pm$ 12.3 & 54.7 \footnotesize$\pm$ 12.2 & 54.9 \footnotesize$\pm$ 12.0 \\
GNN-LRP & 71.7 \footnotesize$\pm$ 7.3 & 78.6  \footnotesize$\pm$ 9.2 & 43.5 \footnotesize$\pm$ 16.7 & 53.4 \footnotesize$\pm$ 17.1 \\
\textit{SubgraphX} & 72.2 \footnotesize$\pm$ 2.1 & 76.1  \footnotesize$\pm$ 2.8 & 47.6 \footnotesize$\pm$ 5.9 & 56.8 \footnotesize$\pm$ 3.3 \\

\midrule
\textsc{L2xGin} & 74.1 \footnotesize$\pm$ 4.3 & 65.6  \footnotesize$\pm$ 3.9 & \textbf{82.8} \footnotesize$\pm$ 5.2 & \textbf{70.7} \footnotesize$\pm$ 4.3 \\
$\textsc{L2xGin}_{dsc}$  & 71.0 \footnotesize$\pm$ 3.0 & 62.4  \footnotesize$\pm$ 4.1 & 78.1 \footnotesize$\pm$ 3.2 & 66.9 \footnotesize$\pm$ 3.5 \\
\bottomrule
\end{tabular}
\end{table}

\vspace{-10pt}

\subsubsection{Qualitative Evaluation of the Explanations}
In Figure \ref{fig:subgraphs}, we present some of the subgraphs identified by \exgnn when combined with two different base GNNs. Based on prior studies and chemical domain knowledge \citep{debnath1991structure,lin2021generative, tan2022learning}, carbon rings (the black circles in the pictures) and $\text{NO}_2$ groups are known to be mutagenic. Interestingly, we can notice that, when using the information of connected subgraphs, the models are able to recognize a complete carbon ring with a $\text{NO}_2$ group in most of the cases. In some cases, the carbon ring is not complete, but the explanations are still helpful to understand which motifs are potentially important for the prediction. With the subscript $dsc$, we can observe the results of the sampling strategy when we do not require subgraphs to be connected. In this case, the carbon rings are not always identified. Instead, the $\text{NO}_2$ group is always considered important for the prediction. More generally, as also reported in \citet{yuan2021subgraphx}, studying connected subgraphs results in more natural motifs compared to the motifs obtained without the connectedness constraint. A visual comparison of the explanations generated by \exgnn and by the baselines can be found in Figure \ref{fig:subgraphs_all} in the appendix.

\begin{table}[h]
\centering
\begin{tabular}{p{1.2cm}CCCCCC}
\toprule

$\textsc{L2xGcn}$  & 
\includegraphics[width=5.5em]{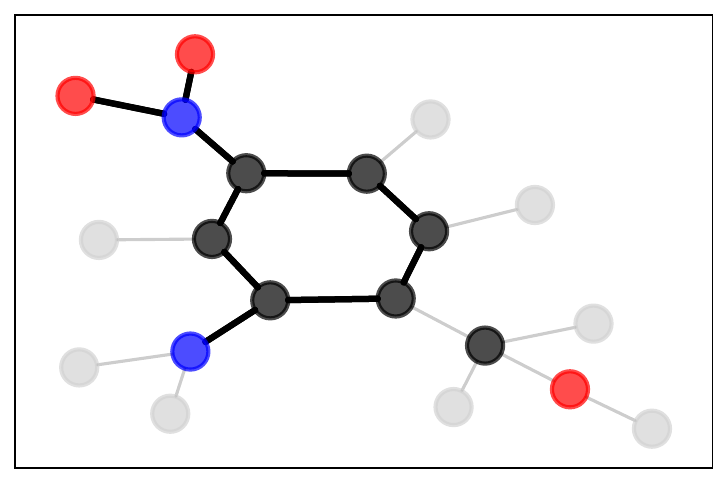} &
\includegraphics[width=5.5em]{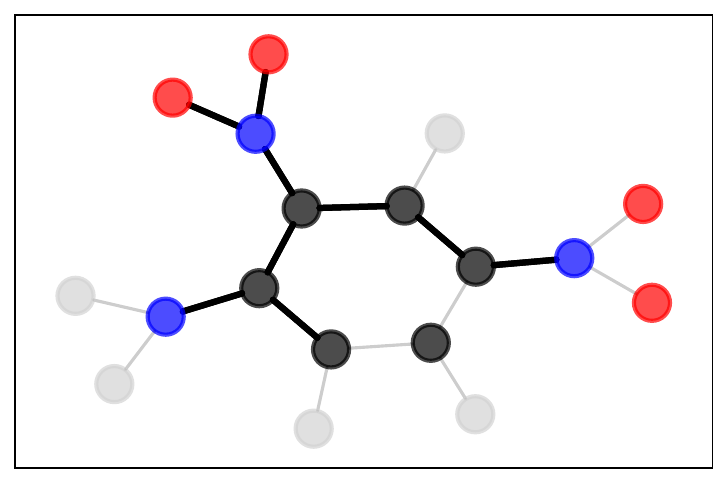} &
\includegraphics[width=5.5em]{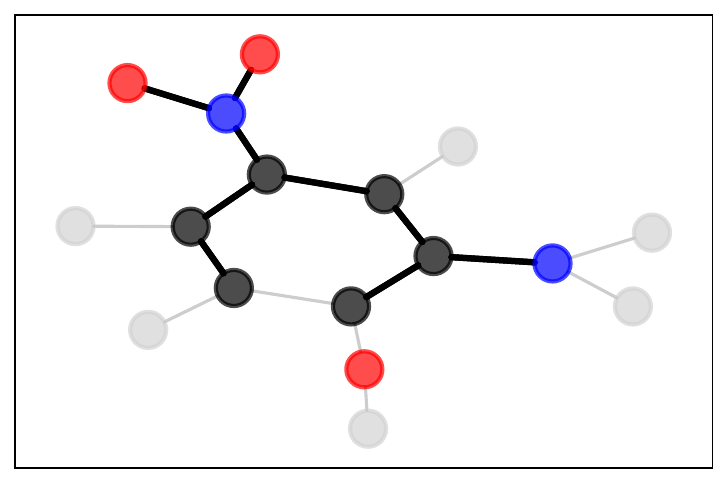}&
\includegraphics[width=5.5em]{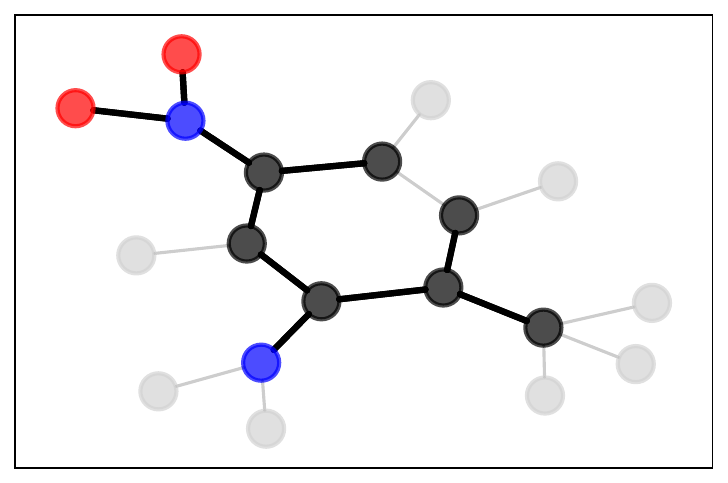}&
\includegraphics[width=5.5em]{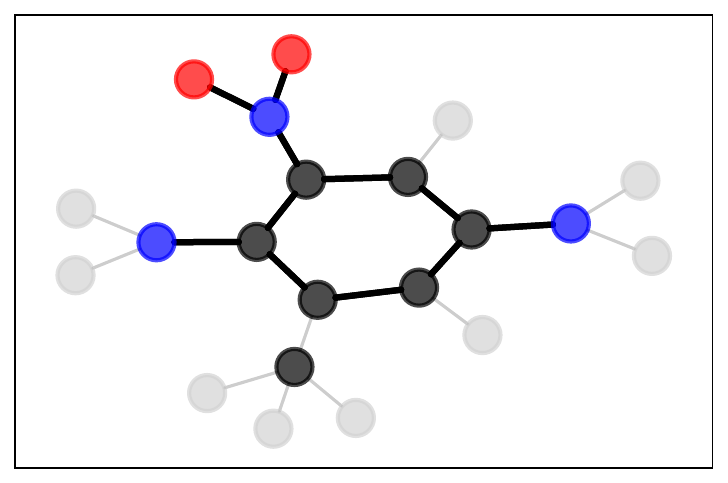}&
\includegraphics[width=5.5em]{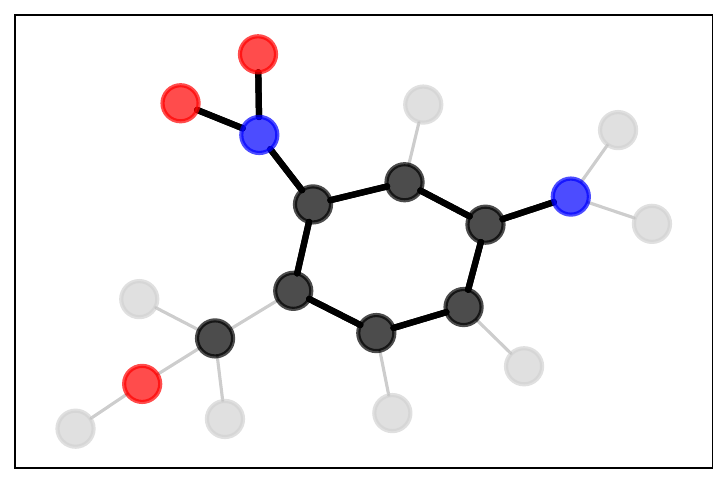}\\

$\textsc{L2xGcn}_{dsc}$  & 
\includegraphics[width=5.5em]{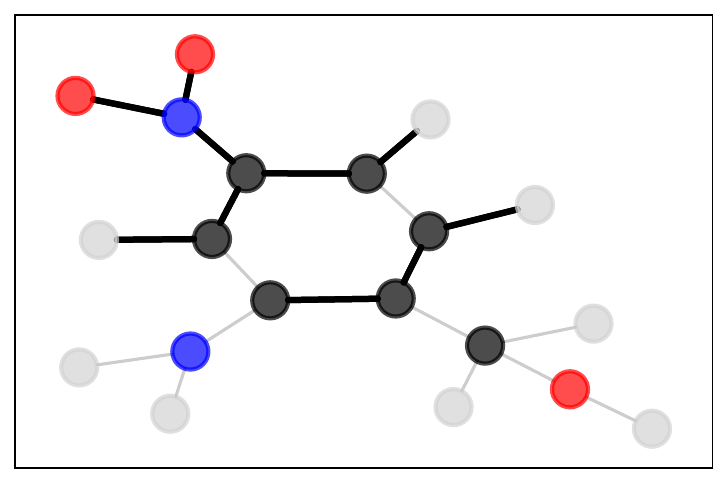} &
\includegraphics[width=5.5em]{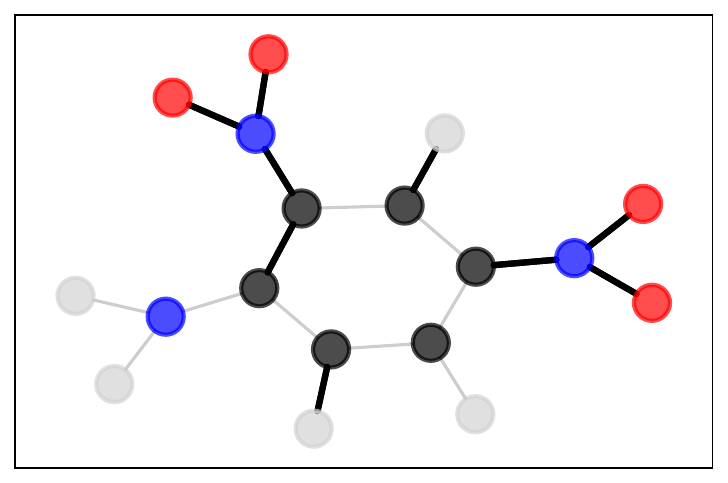} &
\includegraphics[width=5.5em]{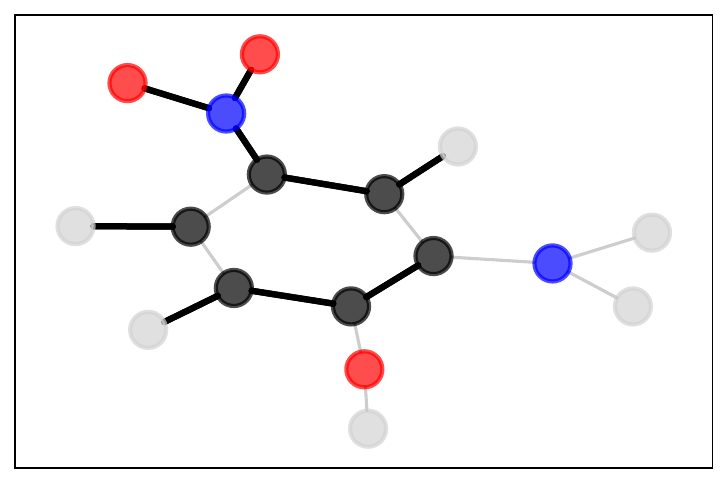} &
\includegraphics[width=5.5em]{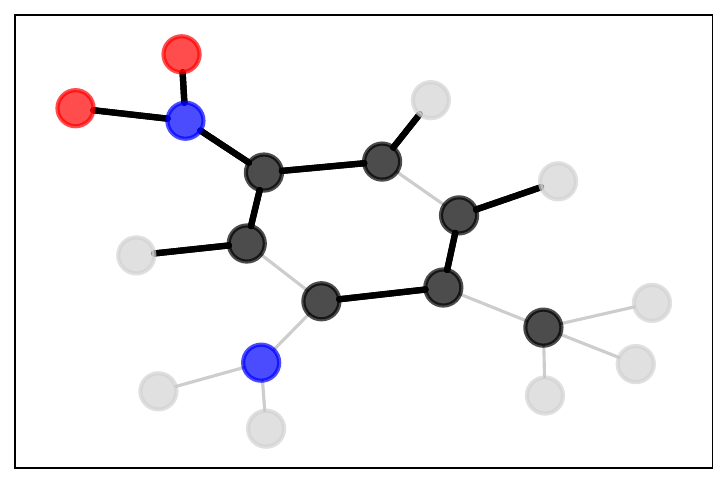} &
\includegraphics[width=5.5em]{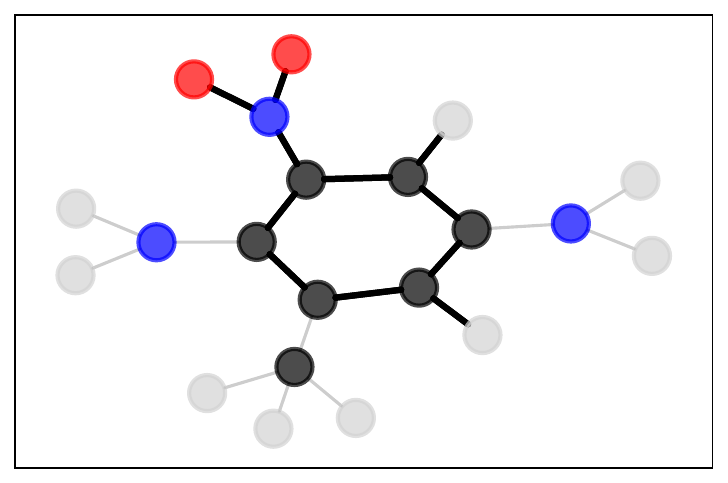} &
\includegraphics[width=5.5em]{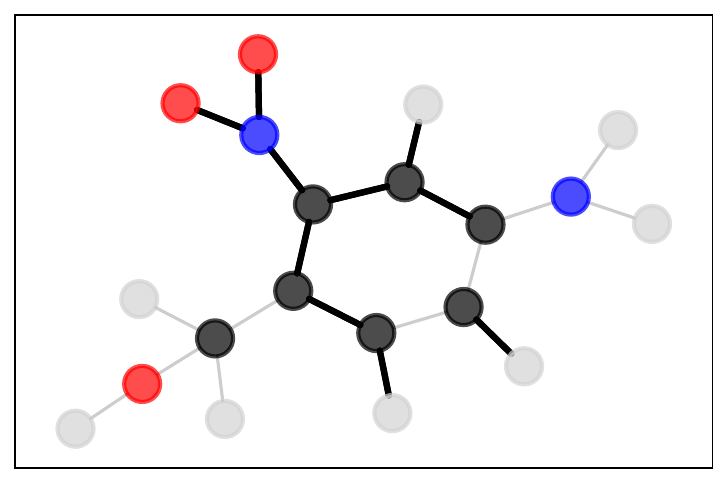}\\

\midrule 

$\textsc{L2xGin}$  & 
\includegraphics[width=5.5em]{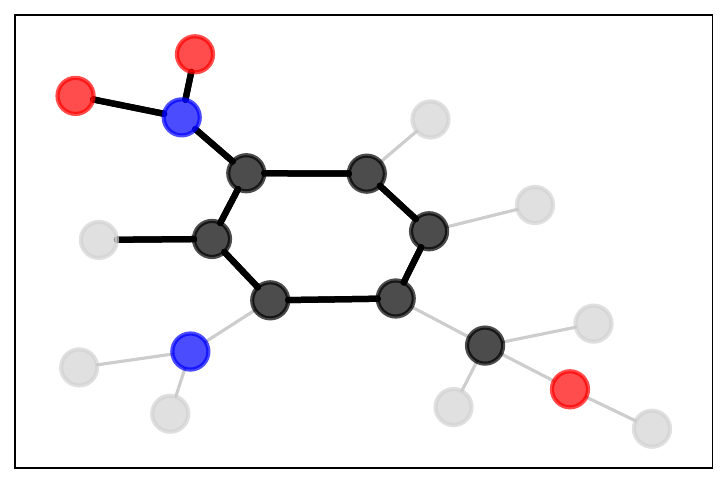} &
\includegraphics[width=5.5em]{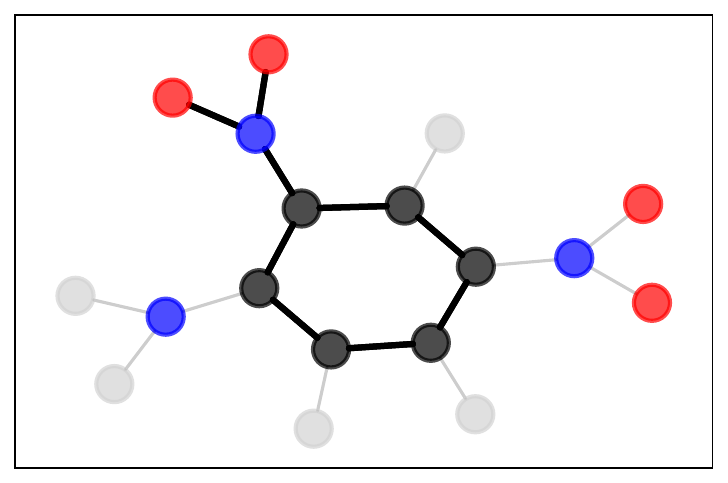}&
\includegraphics[width=5.5em]{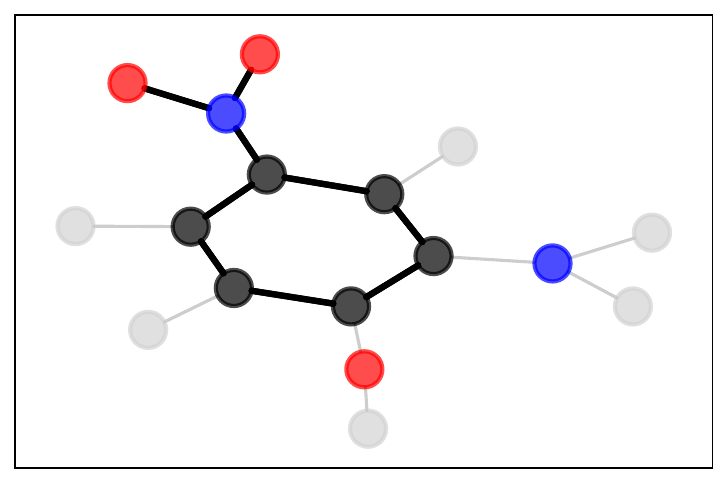}&
\includegraphics[width=5.5em]{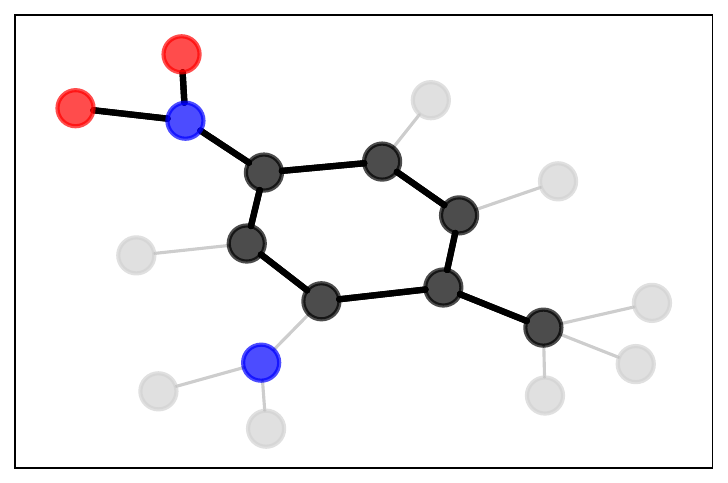}&
\includegraphics[width=5.5em]{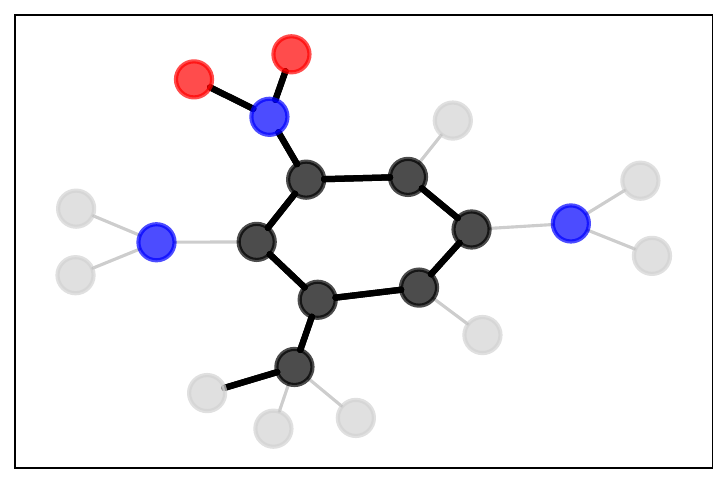}&
\includegraphics[width=5.5em]{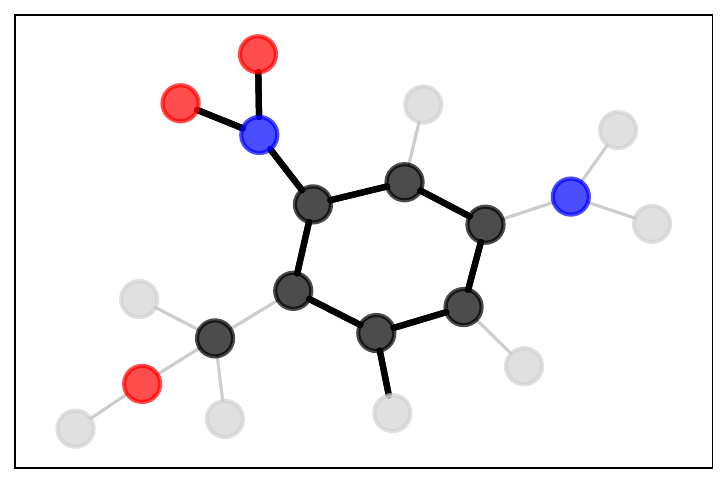}\\

$\textsc{L2xGin}_{dsc}$  & 
\includegraphics[width=5.5em]{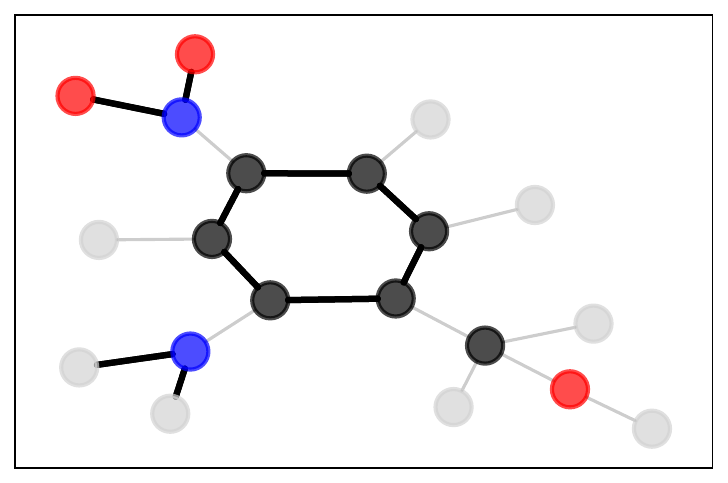} &
\includegraphics[width=5.5em]{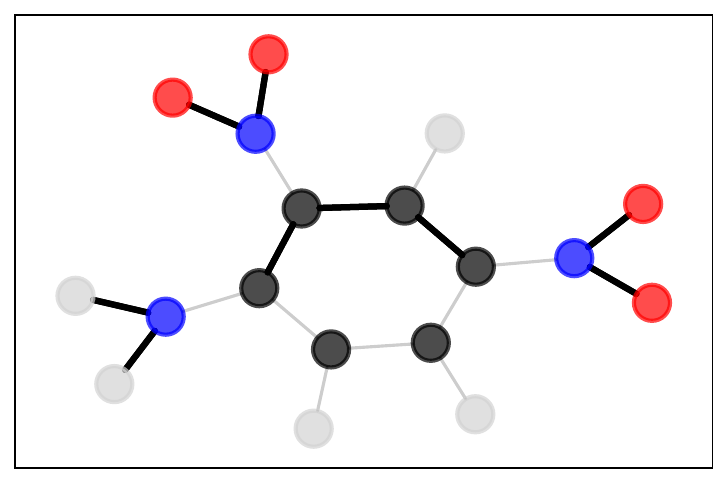} & 
\includegraphics[width=5.5em]{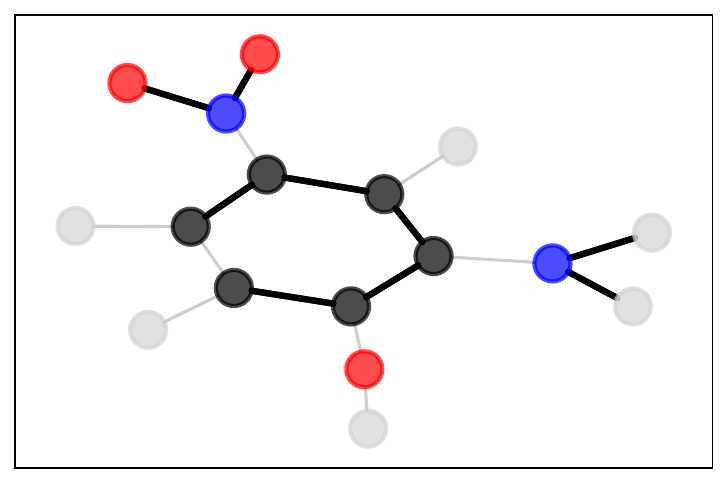} &
\includegraphics[width=5.5em]{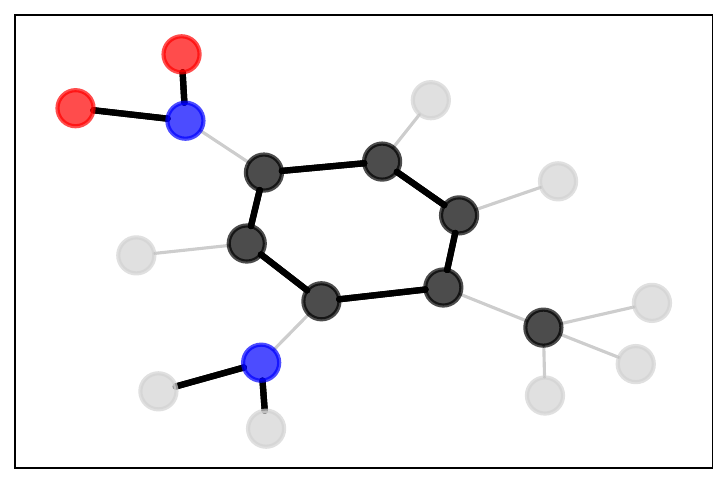} &
\includegraphics[width=5.5em]{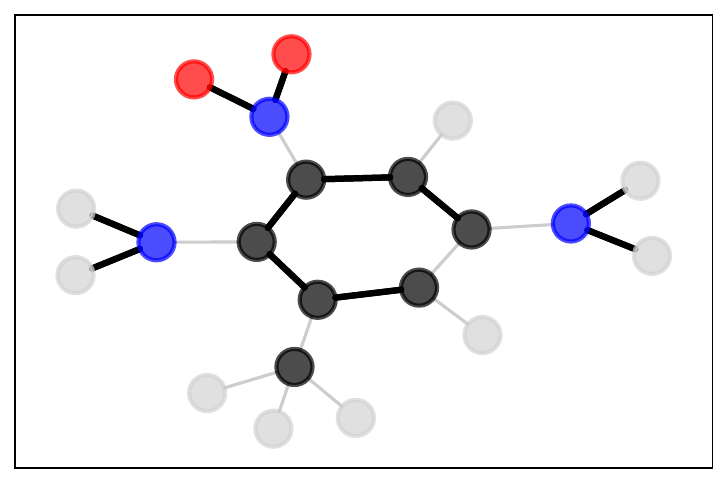} &
\includegraphics[width=5.5em]{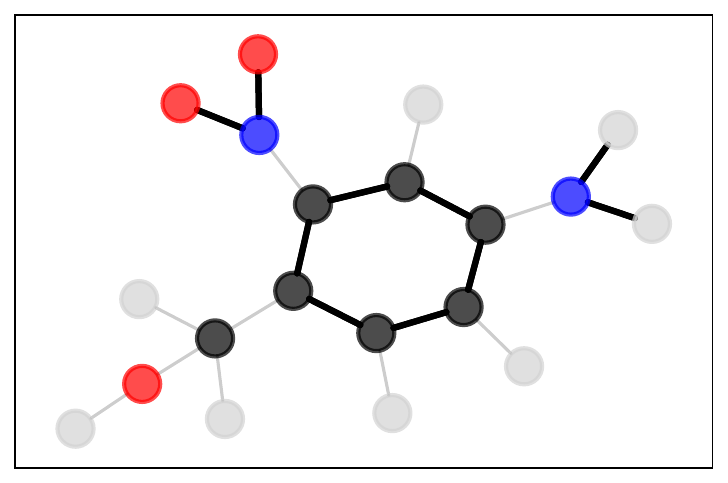}\\ 
\bottomrule 
\end{tabular}
\captionof{figure}{Visualization of some of the subgraphs selected by \exgnn for MUTAG$_0$ on the test set. The solid edges represent the ones sampled by our approach. The subscript \textit{dsc} indicates the maximum weight \textit{k}-edge subgraph problem (i.e., possibly disconnected subgraphs). Black, blue, red, and gray nodes represent carbon (C), nitrogen (N), oxygen (O), and hydrogen (H) atoms respectively.}
\label{fig:subgraphs}

\vspace{-.5cm}
\end{table}

\begin{figure*}[b]
\centering
\includegraphics[width=0.85\textwidth]{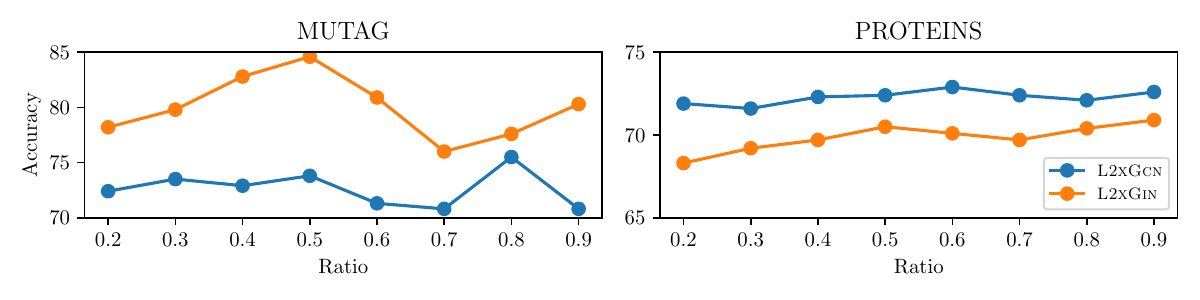} 
\caption{Effect of the edge ratio on the prediction accuracy (\%).}
\label{fig:ratio_plot}
\vspace{-.5cm}
\end{figure*}

\subsubsection{Ablation Study} 
In Section \ref{sec:graph_classification}, we compare the two sampling strategies. From the results, the connected sampling is able to get better results than the non-connected counterpart on most datasets. In fact, the connectivity of subgraphs is essential to grasp the complete information about the important patterns, especially for chemical compound data where connected atoms are usually expected to create molecules or chemical groups. This aspect is also supported by the results obtained in the explanation accuracy task, where the connected strategy returns better explanations for the chemical dataset. Additionally, as previously mentioned, evaluating connected structures rather than just important edges looks more natural and intelligible. In Figure \ref{fig:ratio_plot}, we analyze the effect of the quantity of retained information on the prediction accuracy. A smaller ratio indicates that we retain fewer edges during training and, consequently, the resulting subgraphs are more sparse and, therefore, interpretable. As one can see, this affects the predictive capabilities only when $R$ is small. Starting from $R=0.5$, the ratio does not affect particularly the predictive capabilities of the model. In fact, for graph classification tasks, some of the information contained in the initial computational graph does not condition the prediction as the information may be redundant or noisy. For instance, considering the MUTAG dataset, we know that the initial graphs contain on average 20 edges. The discriminative motif benzene-$\text{NO}_2$, instead, contains around 9 edges, meaning that we ideally need 50\% of the original edges to obtain good results. This is in line with the findings of this analysis and the graph classification results previously reported in Tables \ref{tab:accuracy} and \ref{tab:expl_accuracy}.

    

\subsubsection{Shortcut Learning Detection} 
By generating \textit{faithful} subgraph explanations, our approach can be used to detect whether the predictive model is focusing on the expected features or if it is affected by shortcut learning. This is particularly important for GNNs, where seemingly small implementation differences can influence the learning process of the model \citep{schlichtkrull2021interpreting}. To this end, we use the BA2Motifs dataset \citep{luo2020pgexplainer}. We trained two different models, GCN and GIN, achieving a test accuracy of $0.67$ and $1.0$ respectively. Taking a closer look at the explanations of the first model, we observed that most of the correct predictions were (incorrectly) correlated with the cycle motif and that the explanations were similar to the ones reported in Figure~\ref{fig:shortcut_det}. The explanatory results show that the model is not learning the expected discriminative motifs and, consequently, the accuracy for the test set is poor. This insight can help users to change the configuration of the architecture or to use a different model (e.g., GIN). More generally, the results highlight that faithful explanations can facilitate model analysis and debugging.

\begin{table}[t]
\centering
\begin{tabular}{XXXX}
\toprule
\multicolumn{2}{c}{\textsc{L2xGcn}} & \multicolumn{2}{c}{\textsc{L2xGin}}\\
\midrule
\includegraphics[width=7em]{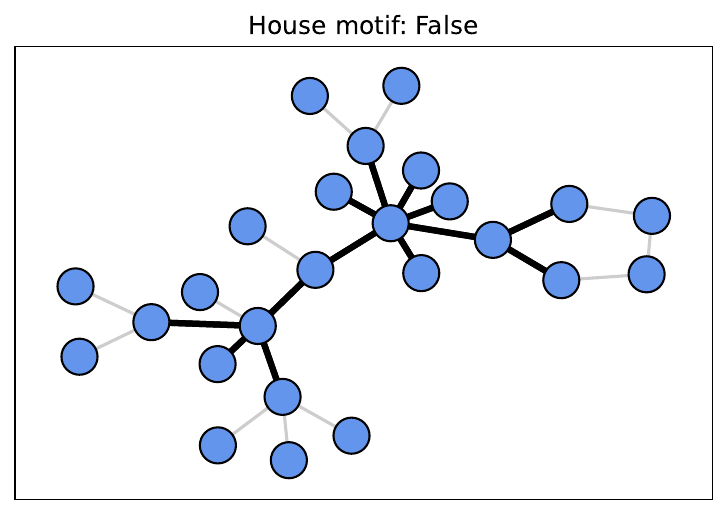} &
\includegraphics[width=7em]{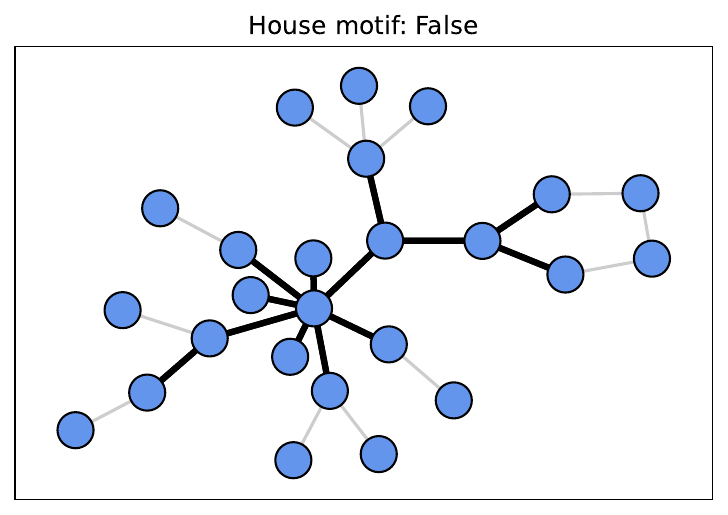} &
\includegraphics[width=7em]{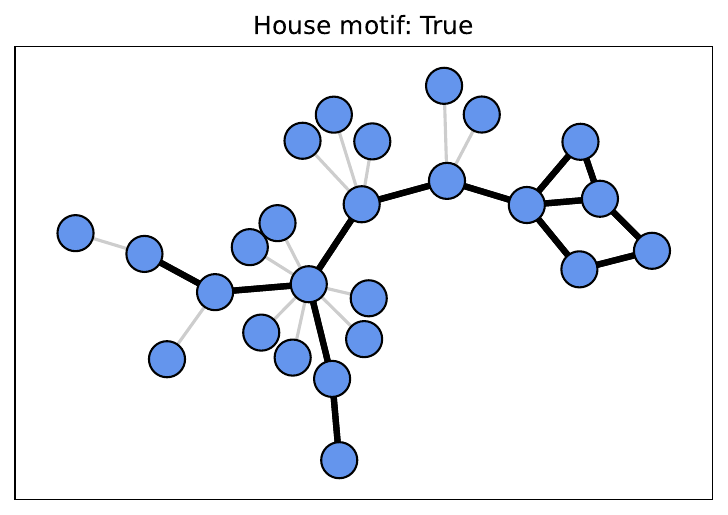} &
\includegraphics[width=7em]{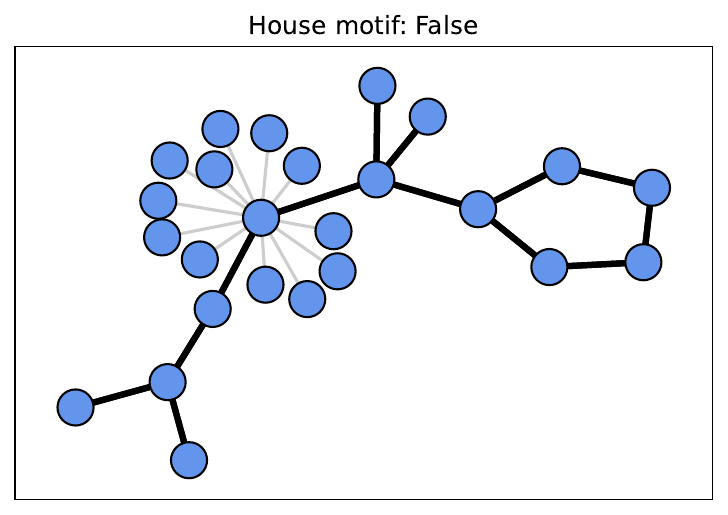} \\
\bottomrule 
\end{tabular}
\captionof{figure}{Example of model reasoning understanding through the visualization of the generated explanations.}
\label{fig:shortcut_det}
\end{table}

\section{Conclusion and Limitations}
We propose \exgnn
, a framework that can be integrated into GNN architectures to learn to generate explanatory subgraphs which are exclusively used for the models' predictions. Our experimental findings demonstrate that the integration of \exgnn with base GNNs does not affect the predictive capabilities of the model for graph classification tasks. Furthermore, according to the definition provided in the paper, the resulting explanations are \textit{faithful} since the retained information is the only one used by the model for prediction. Hence, differently from most of the common techniques, our explanations reveal the rationale of the GNNs and can also be used for model analysis and debugging. A limitation of the approach is the reduced efficiency compared to baseline GNN models. Since we need to integrate an algorithm to compute (approximate) solutions to a combinatorial optimization problem, each forward-pass requires more time and resources. Moreover, depending on the choice of the optimization problem, we might not capture the structure of explanatory motifs required for the application under consideration. 

\section*{Declarations}
Not applicable.


\bibliography{sn-bibliography}

\begin{thebibliography}{53}
\providecommand{\natexlab}[1]{#1}
\providecommand{\url}[1]{{#1}}
\providecommand{\urlprefix}{URL }
\providecommand{\doi}[1]{\url{https://doi.org/#1}}
\providecommand{\eprint}[2][]{\url{#2}}
 \bibcommenthead

\bibitem[{Agarwal et~al(2022{\natexlab{a}})Agarwal, Queen, Lakkaraju, and Zitnik}]{agarwal2022explainable}
Agarwal C, Queen O, Lakkaraju H, et~al (2022{\natexlab{a}}) An explainable ai library for benchmarking graph explainers. In: Workshop on Graph Learning Benchmarks (GLB)

\bibitem[{Agarwal et~al(2022{\natexlab{b}})Agarwal, Zitnik, and Lakkaraju}]{agarwal2022probing}
Agarwal C, Zitnik M, Lakkaraju H (2022{\natexlab{b}}) Probing gnn explainers: A rigorous theoretical and empirical analysis of gnn explanation methods. In: International Conference on Artificial Intelligence and Statistics, pp 8969--8996

\bibitem[{Baldassarre and Azizpour(2019)}]{baldassarre2019explainability}
Baldassarre F, Azizpour H (2019) Explainability techniques for graph convolutional networks. arXiv preprint arXiv:190513686

\bibitem[{Borgwardt et~al(2005)Borgwardt, Ong, Sch{\"o}nauer, Vishwanathan, Smola, and Kriegel}]{borgwardt2005protein}
Borgwardt KM, Ong CS, Sch{\"o}nauer S, et~al (2005) Protein function prediction via graph kernels. Bioinformatics 21(suppl\_1):i47--i56

\bibitem[{Chen et~al(2018)Chen, Song, Wainwright, and Jordan}]{chen2018learning}
Chen J, Song L, Wainwright M, et~al (2018) Learning to explain: An information-theoretic perspective on model interpretation. In: International Conference on Machine Learning, PMLR, pp 883--892

\bibitem[{Debnath et~al(1991)Debnath, Lopez~de Compadre, Debnath, Shusterman, and Hansch}]{debnath1991structure}
Debnath AK, Lopez~de Compadre RL, Debnath G, et~al (1991) Structure-activity relationship of mutagenic aromatic and heteroaromatic nitro compounds. correlation with molecular orbital energies and hydrophobicity. Journal of medicinal chemistry 34(2):786--797

\bibitem[{Dietterich(1998)}]{dietterich1998approximate}
Dietterich TG (1998) Approximate statistical tests for comparing supervised classification learning algorithms. Neural computation 10(7):1895--1923

\bibitem[{Domke(2010)}]{domke2010implicit}
Domke J (2010) Implicit differentiation by perturbation. Advances in Neural Information Processing Systems 23:523--531

\bibitem[{Duval and Malliaros(2021)}]{duval2021graphsvx}
Duval A, Malliaros FD (2021) Graphsvx: Shapley value explanations for graph neural networks. In: Joint European Conference on Machine Learning and Knowledge Discovery in Databases, Springer, pp 302--318

\bibitem[{Faber et~al(2021)Faber, K.~Moghaddam, and Wattenhofer}]{faber2021comparing}
Faber L, K.~Moghaddam A, Wattenhofer R (2021) When comparing to ground truth is wrong: On evaluating gnn explanation methods. In: Proceedings of the 27th ACM SIGKDD Conference on Knowledge Discovery \& Data Mining, pp 332--341

\bibitem[{Feng et~al(2022{\natexlab{a}})Feng, You, Wang, and Tassiulas}]{feng2022kergnns}
Feng A, You C, Wang S, et~al (2022{\natexlab{a}}) Kergnns: Interpretable graph neural networks with graph kernels. In: Proceedings of the AAAI Conference on Artificial Intelligence, pp 6614--6622

\bibitem[{Feng et~al(2022{\natexlab{b}})Feng, Liu, Yang, Tang, Du, and Hu}]{feng2022degree}
Feng Q, Liu N, Yang F, et~al (2022{\natexlab{b}}) {DEGREE}: Decomposition based explanation for graph neural networks. In: International Conference on Learning Representations

\bibitem[{Franceschi et~al(2019)Franceschi, Niepert, Pontil, and He}]{franceschi2019learning}
Franceschi L, Niepert M, Pontil M, et~al (2019) Learning discrete structures for graph neural networks. In: International conference on machine learning, pp 1972--1982

\bibitem[{Gao et~al(2021)Gao, Sun, Bhatt, Yu, Hong, and Zhao}]{gao2021gnes}
Gao Y, Sun T, Bhatt R, et~al (2021) Gnes: Learning to explain graph neural networks. In: 2021 IEEE International Conference on Data Mining (ICDM), IEEE, pp 131--140

\bibitem[{Gui et~al(2022)Gui, Yuan, Wang, Lao, Li, and Ji}]{gui2022flowx}
Gui S, Yuan H, Wang J, et~al (2022) Flowx: Towards explainable graph neural networks via message flows. arXiv preprint arXiv:220612987

\bibitem[{Hamilton et~al(2017)Hamilton, Ying, and Leskovec}]{hamilton2017inductive}
Hamilton WL, Ying R, Leskovec J (2017) Inductive representation learning on large graphs. In: Proceedings of the 31st International Conference on Neural Information Processing Systems, pp 1025--1035

\bibitem[{Huang et~al(2022)Huang, Yamada, Tian, Singh, and Chang}]{huang2020graphlime}
Huang Q, Yamada M, Tian Y, et~al (2022) Graphlime: Local interpretable model explanations for graph neural networks. IEEE Transactions on Knowledge and Data Engineering

\bibitem[{Kipf and Welling(2017)}]{kipf2016semi}
Kipf TN, Welling M (2017) Semi-supervised classification with graph convolutional networks. In: International Conference on Learning Representations

\bibitem[{Lee et~al(2019)Lee, Lee, and Kang}]{lee2019self}
Lee J, Lee I, Kang J (2019) Self-attention graph pooling. In: International conference on machine learning, PMLR, pp 3734--3743

\bibitem[{Lin et~al(2021)Lin, Lan, and Li}]{lin2021generative}
Lin W, Lan H, Li B (2021) Generative causal explanations for graph neural networks. In: International Conference on Machine Learning, PMLR, pp 6666--6679

\bibitem[{Liu et~al(2021)Liu, Luo, Wang, Xie, Yuan, Gui, Yu, Xu, Zhang, Liu, Yan, Liu, Fu, Oztekin, Zhang, and Ji}]{liu2021dig}
Liu M, Luo Y, Wang L, et~al (2021) {DIG}: A turnkey library for diving into graph deep learning research. Journal of Machine Learning Research 22(240):1--9. \urlprefix\url{http://jmlr.org/papers/v22/21-0343.html}

\bibitem[{Lucic et~al(2022)Lucic, Ter~Hoeve, Tolomei, De~Rijke, and Silvestri}]{lucic2022cf}
Lucic A, Ter~Hoeve MA, Tolomei G, et~al (2022) Cf-gnnexplainer: Counterfactual explanations for graph neural networks. In: International Conference on Artificial Intelligence and Statistics, PMLR, pp 4499--4511

\bibitem[{Luo et~al(2020)Luo, Cheng, Xu, Yu, Zong, Chen, and Zhang}]{luo2020pgexplainer}
Luo D, Cheng W, Xu D, et~al (2020) Parameterized explainer for graph neural network. Advances in neural information processing systems 33:19620--19631

\bibitem[{Magister et~al(2021)Magister, Kazhdan, Singh, and Li{\`o}}]{magister2021gcexplainer}
Magister LC, Kazhdan D, Singh V, et~al (2021) Gcexplainer: Human-in-the-loop concept-based explanations for graph neural networks. arXiv preprint arXiv:210711889

\bibitem[{Miao et~al(2022)Miao, Liu, and Li}]{miao2022interpretable}
Miao S, Liu M, Li P (2022) Interpretable and generalizable graph learning via stochastic attention mechanism. In: International Conference on Machine Learning, PMLR, pp 15524--15543

\bibitem[{Niepert et~al(2021)Niepert, Minervini, and Franceschi}]{niepert21imle}
Niepert M, Minervini P, Franceschi L (2021) Implicit {MLE:} backpropagating through discrete exponential family distributions. In: NeurIPS. {PMLR}, Proceedings of Machine Learning Research

\bibitem[{{Papandreou} and {Yuille}(2011)}]{Papandreou:2011}
{Papandreou} G, {Yuille} AL (2011) Perturb-and-map random fields: Using discrete optimization to learn and sample from energy models. In: 2011 International Conference on Computer Vision, pp 193--200

\bibitem[{Perotti et~al(2023)Perotti, Bajardi, Bonchi, and Panisson}]{perotti2022graphshap}
Perotti A, Bajardi P, Bonchi F, et~al (2023) Explaining identity-aware graph classifiers through the language of motifs. In: 2023 International Joint Conference on Neural Networks (IJCNN), IEEE, pp 1--8

\bibitem[{Pope et~al(2019)Pope, Kolouri, Rostami, Martin, and Hoffmann}]{pope2019gradcam}
Pope PE, Kolouri S, Rostami M, et~al (2019) Explainability methods for graph convolutional neural networks. In: Proceedings of the IEEE/CVF Conference on Computer Vision and Pattern Recognition, pp 10772--10781

\bibitem[{Qian et~al(2022)Qian, Rattan, Geerts, Niepert, and Morris}]{qian2022ordered}
Qian C, Rattan G, Geerts F, et~al (2022) Ordered subgraph aggregation networks. Advances in Neural Information Processing Systems 35:21030--21045

\bibitem[{Rossi and Ahmed(2015)}]{rossi2015network}
Rossi R, Ahmed N (2015) The network data repository with interactive graph analytics and visualization. In: Twenty-ninth AAAI conference on artificial intelligence

\bibitem[{Rudin(2018)}]{rudin2018stop}
Rudin C (2018) Stop explaining black box machine learning models for high stakes decisions and use interpretable models instead. In: Proceedings of NeurIPS 2018 Workshop on Critiquing and Correcting Trends in Learning

\bibitem[{Sanchez-Lengeling et~al(2020)Sanchez-Lengeling, Wei, Lee, Reif, Wang, Qian, McCloskey, Colwell, and Wiltschko}]{sanchez2020evaluating}
Sanchez-Lengeling B, Wei J, Lee B, et~al (2020) Evaluating attribution for graph neural networks. Advances in neural information processing systems 33:5898--5910

\bibitem[{Schlichtkrull et~al(2021)Schlichtkrull, Cao, and Titov}]{schlichtkrull2021interpreting}
Schlichtkrull MS, Cao ND, Titov I (2021) Interpreting graph neural networks for {\{}nlp{\}} with differentiable edge masking. In: International Conference on Learning Representations

\bibitem[{Schnake et~al(2021)Schnake, Eberle, Lederer, Nakajima, Sch{\"u}tt, M{\"u}ller, and Montavon}]{schnake2020gnnlrp}
Schnake T, Eberle O, Lederer J, et~al (2021) Higher-order explanations of graph neural networks via relevant walks. IEEE transactions on pattern analysis and machine intelligence 44(11):7581--7596

\bibitem[{Schwarzenberg et~al(2019)Schwarzenberg, H{\"u}bner, Harbecke, Alt, and Hennig}]{schwarzenberg2019layerwise}
Schwarzenberg R, H{\"u}bner M, Harbecke D, et~al (2019) Layerwise relevance visualization in convolutional text graph classifiers. arXiv preprint arXiv:190910911

\bibitem[{Tan et~al(2022)Tan, Geng, Fu, Ge, Xu, Li, and Zhang}]{tan2022learning}
Tan J, Geng S, Fu Z, et~al (2022) Learning and evaluating graph neural network explanations based on counterfactual and factual reasoning. In: Proceedings of the ACM Web Conference 2022, pp 1018--1027

\bibitem[{Veli{\v{c}}kovi{\'c} et~al(2018)Veli{\v{c}}kovi{\'c}, Cucurull, Casanova, Romero, Li{\`o}, and Bengio}]{velivckovic2018graph}
Veli{\v{c}}kovi{\'c} P, Cucurull G, Casanova A, et~al (2018) Graph attention networks. In: International Conference on Learning Representations

\bibitem[{Vu and Thai(2020)}]{vu2020pgm}
Vu M, Thai MT (2020) Pgm-explainer: Probabilistic graphical model explanations for graph neural networks. Advances in neural information processing systems 33:12225--12235

\bibitem[{Xu et~al(2018)Xu, Hu, Leskovec, and Jegelka}]{xu2018powerful}
Xu K, Hu W, Leskovec J, et~al (2018) How powerful are graph neural networks? In: International Conference on Learning Representations

\bibitem[{Xuanyuan et~al(2023)Xuanyuan, Barbiero, Georgiev, Magister, and Li{\`o}}]{xuanyuan2022global}
Xuanyuan H, Barbiero P, Georgiev D, et~al (2023) Global concept-based interpretability for graph neural networks via neuron analysis. In: Proceedings of the AAAI Conference on Artificial Intelligence, pp 10675--10683

\bibitem[{Yan et~al(2008)Yan, Cheng, Han, and Yu}]{yan2008mining}
Yan X, Cheng H, Han J, et~al (2008) Mining significant graph patterns by leap search. In: Proceedings of the 2008 ACM SIGMOD international conference on Management of data, pp 433--444

\bibitem[{Yanardag and Vishwanathan(2015)}]{yanardag2015deep}
Yanardag P, Vishwanathan S (2015) Deep graph kernels. In: Proceedings of the 21th ACM SIGKDD international conference on knowledge discovery and data mining, pp 1365--1374

\bibitem[{Ying et~al(2019)Ying, Bourgeois, You, Zitnik, and Leskovec}]{ying2019gnnexplainer}
Ying R, Bourgeois D, You J, et~al (2019) Gnnexplainer: Generating explanations for graph neural networks. Advances in neural information processing systems 32:9240

\bibitem[{Ying et~al(2018)Ying, You, Morris, Ren, Hamilton, and Leskovec}]{ying2018hierarchical}
Ying Z, You J, Morris C, et~al (2018) Hierarchical graph representation learning with differentiable pooling. Advances in neural information processing systems 31

\bibitem[{Yu et~al(2020)Yu, Xu, Rong, Bian, Huang, and He}]{yu2020graph}
Yu J, Xu T, Rong Y, et~al (2020) Graph information bottleneck for subgraph recognition. In: International Conference on Learning Representations

\bibitem[{Yu and Gao(2022)}]{yu2022motifexplainer}
Yu Z, Gao H (2022) Motifexplainer: a motif-based graph neural network explainer. arXiv preprint arXiv:220200519

\bibitem[{Yuan et~al(2020)Yuan, Tang, Hu, and Ji}]{yuan2020xgnn}
Yuan H, Tang J, Hu X, et~al (2020) Xgnn: Towards model-level explanations of graph neural networks. In: Proceedings of the 26th ACM SIGKDD International Conference on Knowledge Discovery \& Data Mining, pp 430--438

\bibitem[{Yuan et~al(2021)Yuan, Yu, Wang, Li, and Ji}]{yuan2021subgraphx}
Yuan H, Yu H, Wang J, et~al (2021) On explainability of graph neural networks via subgraph explorations. In: International conference on machine learning, PMLR, pp 12241--12252

\bibitem[{Yuan et~al(2022)Yuan, Yu, Gui, and Ji}]{yuan2020explainability}
Yuan H, Yu H, Gui S, et~al (2022) Explainability in graph neural networks: A taxonomic survey. IEEE transactions on pattern analysis and machine intelligence 45(5):5782--5799

\bibitem[{Zhang et~al(2018)Zhang, Cui, Neumann, and Chen}]{zhang2018end}
Zhang M, Cui Z, Neumann M, et~al (2018) An end-to-end deep learning architecture for graph classification. In: Proceedings of the AAAI conference on artificial intelligence

\bibitem[{Zhang et~al(2022)Zhang, Liu, Wang, Lu, and Lee}]{zhang2021protgnn}
Zhang Z, Liu Q, Wang H, et~al (2022) Protgnn: Towards self-explaining graph neural networks. In: Proceedings of the AAAI Conference on Artificial Intelligence, pp 9127--9135

\bibitem[{Zhu et~al(2021)Zhu, Xu, Zhang, Liu, Wu, and Wang}]{zhu2021deep}
Zhu Y, Xu W, Zhang J, et~al (2021) Deep graph structure learning for robust representations: A survey. arXiv preprint arXiv:210303036

\end{thebibliography}

\begin{appendices}

\section{Additional Results}\label{secA1}

\subsection{Graph Classification Accuracy for Synthetic Datasets}
In Table \ref{tab:test_accuracy}, we report the graph classification accuracy for the datasets used in our explanation accuracy experiment. In particular, we compare the 3-layer GIN architecture used for generating explanations through post-hoc techniques and the same architecture when our approach is integrated during training. The data splits are the same used in the previous evaluation. Again, the results demonstrate that the integration of \exgnn does not degrade significantly the predictive capabilities of the original model.

\begin{table}[h]
\caption{Comparison of the prediction test accuracy (\%) for synthetic graph classification tasks between a 3-layer GIN architecture and the same architecture with our approach integrated.}
\label{tab:test_accuracy}
\centering
\begin{tabular}{lcc}
\toprule
Dataset & BA-2MOTIFS & MUTAG$_0$ \\
\midrule
GIN           & 100.0  \footnotesize$\pm$ 0.00 & 
                         100.0  \footnotesize$\pm$ 0.00 \\
$\textsc{L2xGin}_{dsc}$  & 99.6  \footnotesize$\pm$ 0.04 & 
                         99.9  \footnotesize$\pm$ 0.01 \\
$\textsc{L2xGin}$ &      99.6  \footnotesize$\pm$ 0.04 & 
                         99.6  \footnotesize$\pm$ 0.03 \\
\bottomrule
\end{tabular}
\end{table}

\subsection{Explanation Consistency}
One crucial property for explanatory methods is consistency. For instance, if an explanation algorithm is applied to the same data instance multiple times, the generated explanations should be unchanged. Also, when different random seeds are used for the same architecture, the generated explanations should be stable. For the first case, we report the results in Table \ref{tab:expl_consistency_test}. Our method preserves its consistency when it is applied to the same data instance multiple times at test time. This is in line with the assumptions of our approach. In fact, since perturbations for subgraph sampling are removed at test time, this behavior is guaranteed. In Table \ref{tab:expl_consistency}, we report the explanation accuracy of our approach when using the same backbone architecture with different model initializations. Specifically, we compare a 3-layer GIN model using five different seeds for model initialization on the same data split. From the results, we can observe the ability of our method to generate consistent explanations irrespective of the differences across models.
\pagebreak

\begin{table}[h]
    \caption{Explanation accuracy (\%) on multiple test runs over the same data instances.}
    \label{tab:expl_consistency_test}
    \centering
    \begin{tabular}{lcccccccc}
    \toprule
    \multicolumn{1}{l}{Dataset} & \multicolumn{4}{c}{BA-2MOTIFS} \\ \cmidrule(r){2-5}
    \multicolumn{1}{c}{}   & Acc.    & Pr.   & Rec. & F$_1$ \\
    
    \midrule
    $\textsc{L2xGin}$  & 75.9  \footnotesize$\pm$ 0.0 & 47.0  \footnotesize$\pm$ 0.0 & 90.0  \footnotesize$\pm$ 0.0 & 61.7  \footnotesize$\pm$ 0.0\\
    $\textsc{L2xGin}_{dsc}$   & 77.9  \footnotesize$\pm$ 0.0 & 49.5  \footnotesize$\pm$ 0.0 & 94.4  \footnotesize$\pm$ 0.0 & 64.8  \footnotesize$\pm$ 0.0 \\
    \midrule
    
    \multicolumn{1}{l}{Dataset}& \multicolumn{4}{c}{MUTAG$_{0}$}\\ \cmidrule(r){2-5}
    \multicolumn{1}{c}{}   & Acc.    & Pr.   & Rec. & F$_1$ \\
    
    \midrule
    $\textsc{L2xGin}$ & 71.0 \footnotesize$\pm$ 0.0 & 63.7  \footnotesize$\pm$ 0.0 & 78.4 \footnotesize$\pm$ 0.0 & 67.7 \footnotesize$\pm$ 0.0 \\
    $\textsc{L2xGin}_{dsc}$ & 70.8 \footnotesize$\pm$ 0.0 & 63.4  \footnotesize$\pm$ 0.0 & 77.0 \footnotesize$\pm$ 0.0 & 67.1 \footnotesize$\pm$ 0.0 \\
    \bottomrule
    
    \end{tabular}
    \end{table}

\begin{table}[h]
\caption{Explanation accuracy (\%) on different model initializations using a 3-layer GIN architecture.}
\label{tab:expl_consistency}
\centering
\begin{tabular}{lcccccccc}
\toprule
\multicolumn{1}{l}{Dataset} & \multicolumn{4}{c}{BA-2MOTIFS}\\ \cmidrule(r){2-5}
\multicolumn{1}{c}{}   & Acc.    & Pr.   & Rec. & F$_1$ \\

\midrule
$\textsc{L2xGin}$  & 75.9  \footnotesize$\pm$ 0.0 & 47.0  \footnotesize$\pm$ 0.0 & 90.0  \footnotesize$\pm$ 0.0 & 61.7  \footnotesize$\pm$ 0.0 \\
$\textsc{L2xGin}_{dsc}$   & 77.9  \footnotesize$\pm$ 0.0 & 49.4  \footnotesize$\pm$ 0.0 & 94.3  \footnotesize$\pm$ 0.1 & 64.8  \footnotesize$\pm$ 0.0 \\
\midrule

\multicolumn{1}{l}{Dataset} & \multicolumn{4}{c}{MUTAG$_{0}$}\\ \cmidrule(r){2-5}
\multicolumn{1}{c}{}   & Acc.    & Pr.   & Rec. & F$_1$ \\

\midrule
$\textsc{L2xGin}$  & 73.8 \footnotesize$\pm$ 3.8 & 66.8  \footnotesize$\pm$ 3.7 & 81.8 \footnotesize$\pm$ 4.2 & 71.2 \footnotesize$\pm$ 3.8 \\
$\textsc{L2xGin}_{dsc}$   & 68.7 \footnotesize$\pm$ 1.6 & 61.5  \footnotesize$\pm$ 2.6 & 74.4 \footnotesize$\pm$ 3.7 & 64.9 \footnotesize$\pm$ 1.9 \\
\bottomrule
\end{tabular}
\end{table}

\subsection{Time Complexity Analysis}
As reported in \citet{feng2022kergnns}, most message-passing architectures have a time complexity of $\mathcal{O}(n^2)$. Thus, since \exgnn uses native GNNs, it also has a worst-case complexity of  $\mathcal{O}(n^2)$. KerGNN instead, being based on graph kernels, has a time complexity which varies between $\mathcal{O}(n^2)$ and $\mathcal{O}(n^3)$ in case the graph is fully-connected. The time overhead of $\optalgo$ depends on the algorithm used. For the maximum-weight $k$-edge subgraph problem, we only need to select the $k$ edges with the maximum weights. To find the maximum-weight $k$-edge \textit{connected} subgraph, the greedy algorithm needs to find the maximum weight edge among all edges adjacent to the currently selected subgraph which, in the worst case, are all edges. In both cases, this results in a longer training time compared to the original architecture. To conclude the comparison with the considered post-hoc techniques, we analyze the time efficiency for generating explanations for unseen data. For a coherent comparison, we use the same test splits used for the previous experiments. Table \ref{tab:test_exec} reports the average time cost to obtain the explanations related to the graphs in the test sets. From the results, we can notice that the training speed of \textsc{L2xGin} is indeed slower with respect to the original implementation ($\sim$1s training time). However, considering the results for generating explanations at test time, our method results in much faster computation time than most of the compared baselines. As such, if we consider strong baselines like SubgraphX or GNN-LRP, the combined computation time is in our favor. 

\begin{table}[h]
\caption{Evaluation of the average execution time (in sec) for generating explanations of unseen data. In brackets, we report the average training time for our approach.}
\label{tab:test_exec}
\centering
\begin{tabular}{lcccc}
\toprule
\multicolumn{1}{l}{Dataset} & \multicolumn{2}{c}{BA-2MOTIFS (size = 100)} & \multicolumn{2}{c}{MUTAG$_{0}$ (size = 223)}\\ \cmidrule(r){2-3}\cmidrule(r){4-5} 
\multicolumn{1}{c}{}   & Nodes (avg)  &  Edges (avg)  & Nodes (avg)  &  Edges (avg) \\
&  25.0   & 51.0 &  31.74   & 32.54 \\
\midrule
GNN-Exp. & \multicolumn{2}{c}{227.10  \footnotesize$\pm$ 1.18} & 
            \multicolumn{2}{c}{529.43  \footnotesize$\pm$ 2.02} \\
GradCAM & \multicolumn{2}{c}{1.41  \footnotesize$\pm$ 0.17} & 
            \multicolumn{2}{c}{2.66  \footnotesize$\pm$ 0.03} \\
PGE-Exp. & \multicolumn{2}{c}{0.64  \footnotesize$\pm$ 0.02} & 
            \multicolumn{2}{c}{1.74  \footnotesize$\pm$ 0.25} \\
GNN-LRP & \multicolumn{2}{c}{918.42  \footnotesize$\pm$ 32.26} & 
            \multicolumn{2}{c}{1938.98  \footnotesize$\pm$ 38.14} \\
\textit{SubgraphX} & \multicolumn{2}{c}{3500.66  \footnotesize$\pm$ 358.31} & 
            \multicolumn{2}{c}{25585.26  \footnotesize$\pm$ 1143.18} \\

\midrule
$\textsc{L2xGin}_{dsc}$  (+19.7s) & \multicolumn{2}{c}{0.74  \footnotesize$\pm$ 0.04} & 
            \multicolumn{2}{c}{0.61  \footnotesize$\pm$ 0.01} \\
\textsc{L2xGin} \;\;\;\, (+51.4s) & \multicolumn{2}{c}{0.73  \footnotesize$\pm$ 0.02} & 
            \multicolumn{2}{c}{1.08  \footnotesize$\pm$ 0.02} \\
\bottomrule
\end{tabular}
\end{table}

\subsection{Visual Comparison of Generated Explanations}
In Figure \ref{fig:subgraphs_all} we provide a visual analysis of the explanations generated by the methods considered in our experiments. The graph visualization supports the numerical evaluation carried on in the main paper. In fact, one can see that the explanations generated by the post-hoc approaches may vary substantially depending on the given input graph. In our case, instead, the explanations remain constant regardless of the input information. This claim supports the explanation accuracy analysis, where our approach has one of the smallest standard deviation among all the considered methods. Additionally, we also included the explanations generated with an attention-based GNN, namely GAT \citep{velivckovic2018graph}. Although having a similar predictive performance in the graph classification task (99.6 $\pm$ 0.03), the resulting explanations are not qualitatively comparable with our approach. This is in line with previous works \citep{ying2019gnnexplainer, yu2020graph, miao2022interpretable} asserting that graph attention models are not able to generate attention weights with high-fidelity, and consequently, cannot provide faithful and meaningful explanations.

\begin{table}[b]
\centering
\begin{tabular}{p{1.2cm}CCCCCC}
\toprule

GNN-Exp.  & 
\includegraphics[width=5.5em]{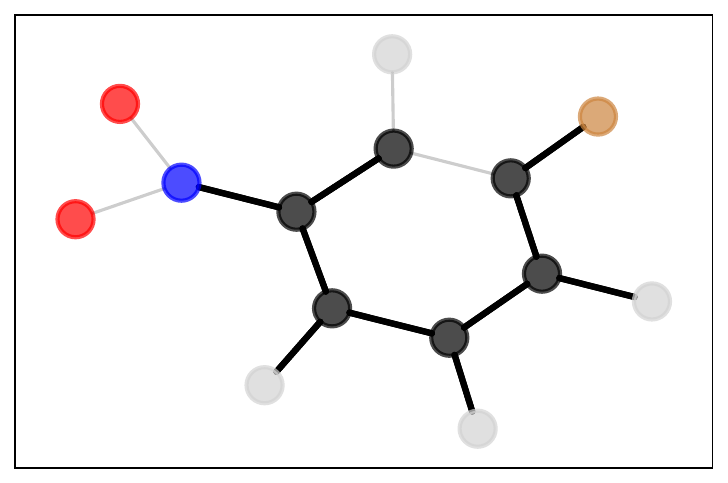}&
\includegraphics[width=5.5em]{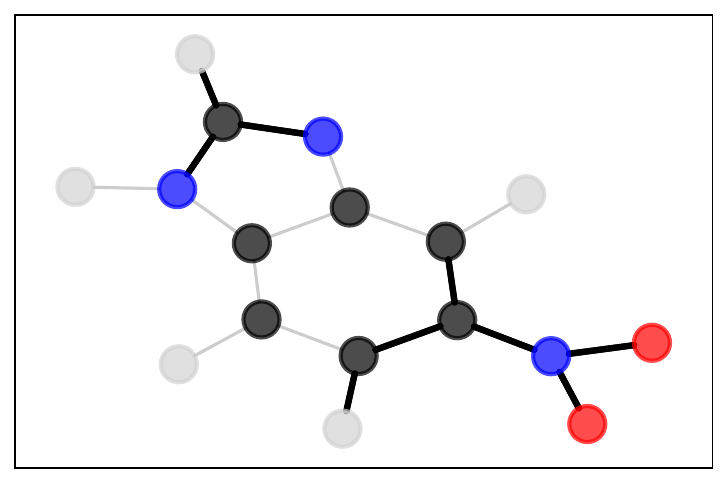} &
\includegraphics[width=5.5em]{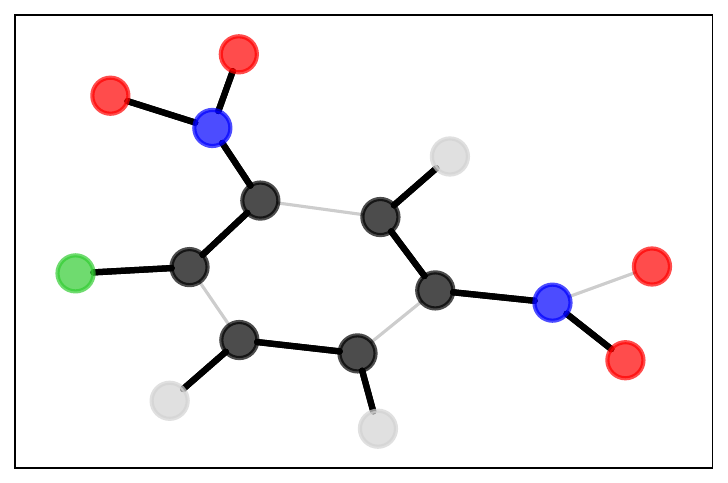}& 
\includegraphics[width=5.5em]{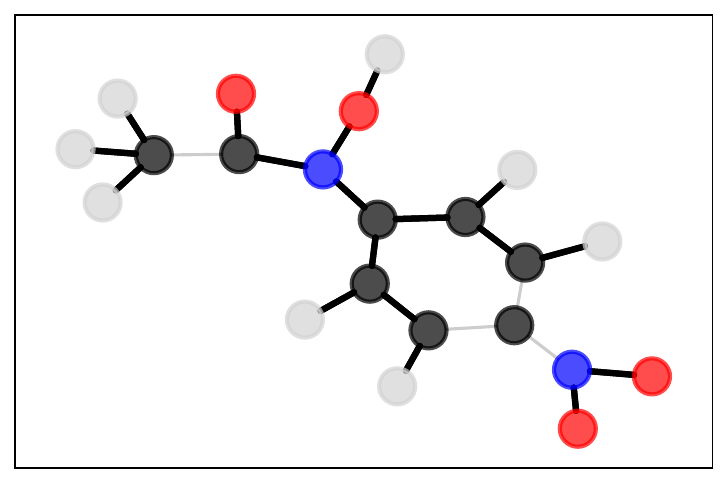}&
\includegraphics[width=5.5em]{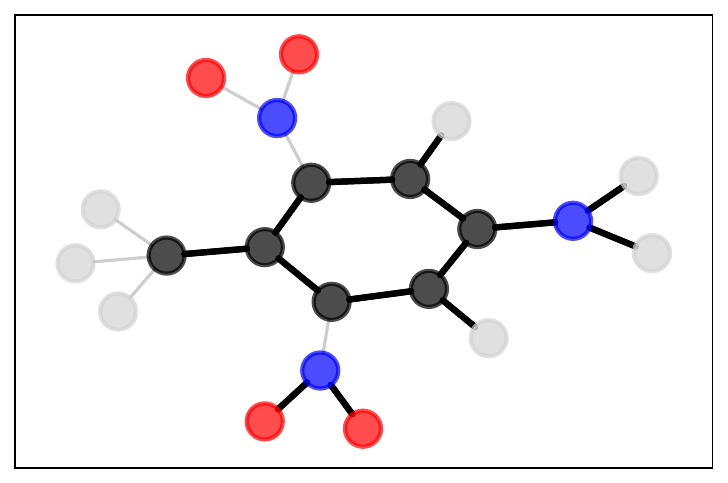} &
\includegraphics[width=5.5em]{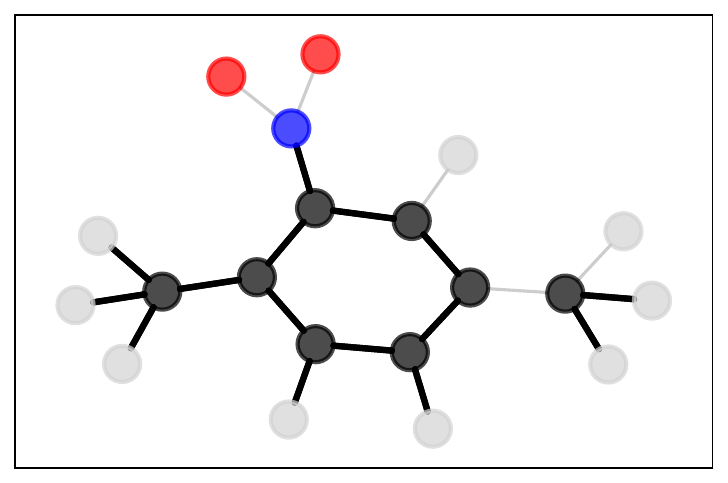}\\ 

\midrule 

GradCAM  & 
\includegraphics[width=5.5em]{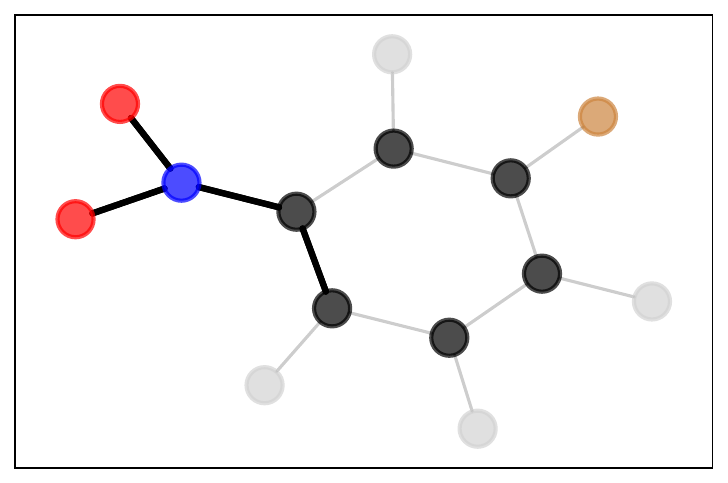}&
\includegraphics[width=5.5em]{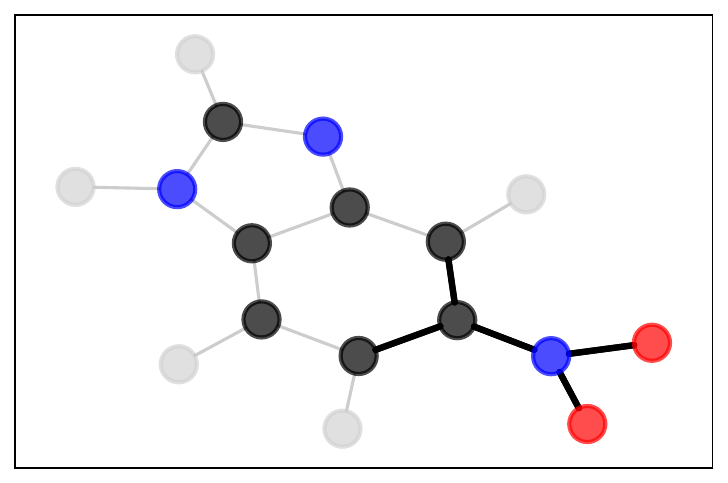}&
\includegraphics[width=5.5em]{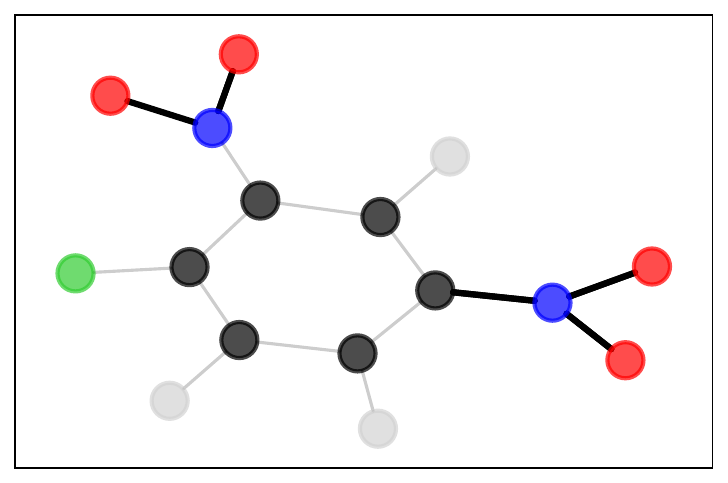}&
\includegraphics[width=5.5em]{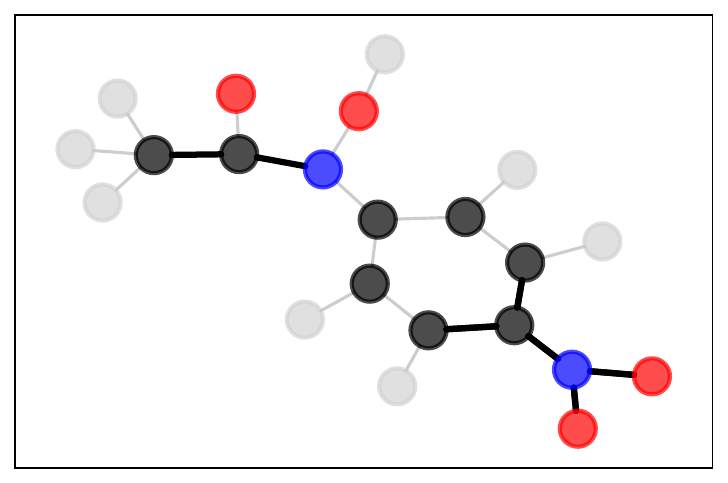}&
\includegraphics[width=5.5em]{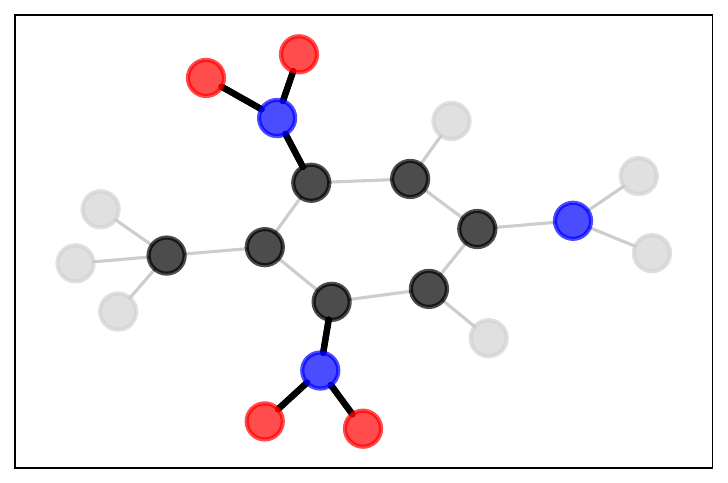}&
\includegraphics[width=5.5em]{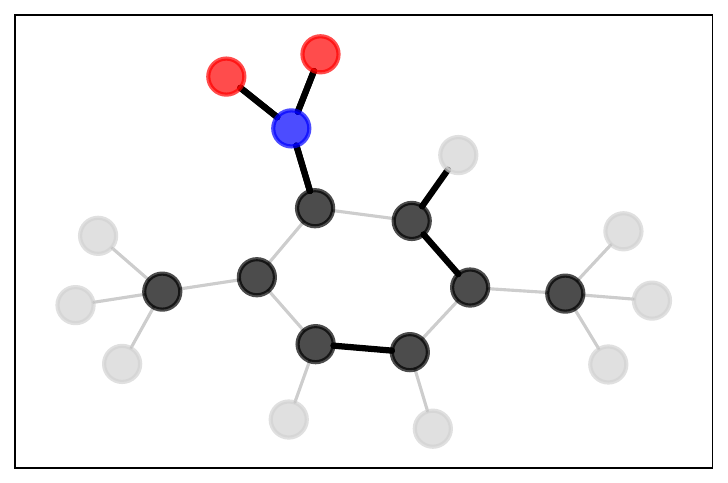} \\

\midrule 

PGE-Exp.  & 
\includegraphics[width=5.5em]{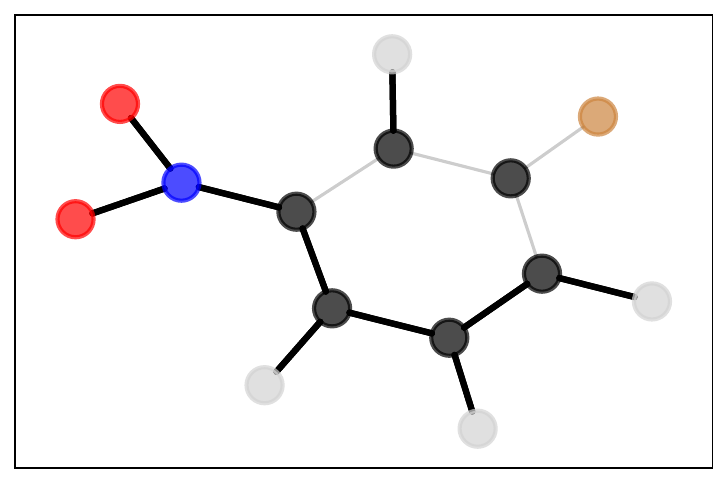}&
\includegraphics[width=5.5em]{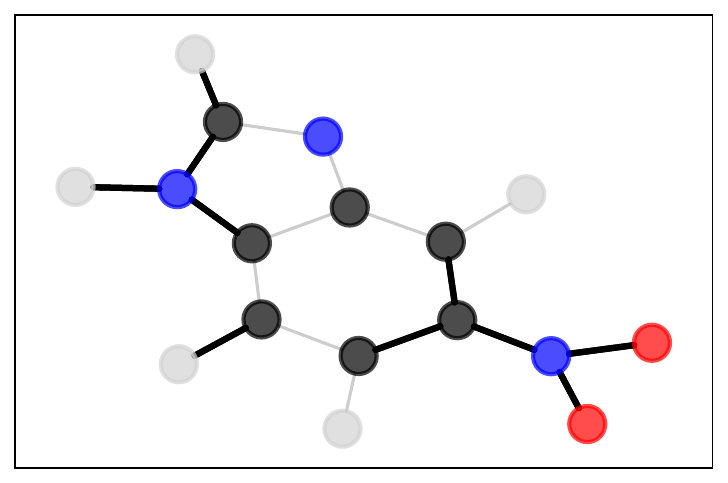}&
\includegraphics[width=5.5em]{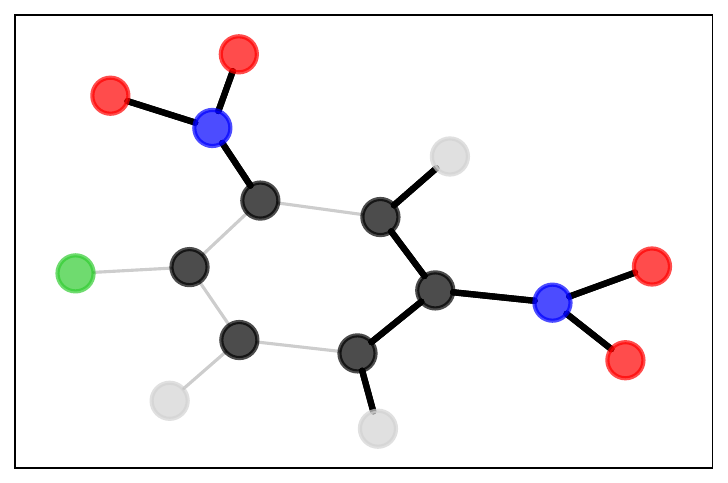}&
\includegraphics[width=5.5em]{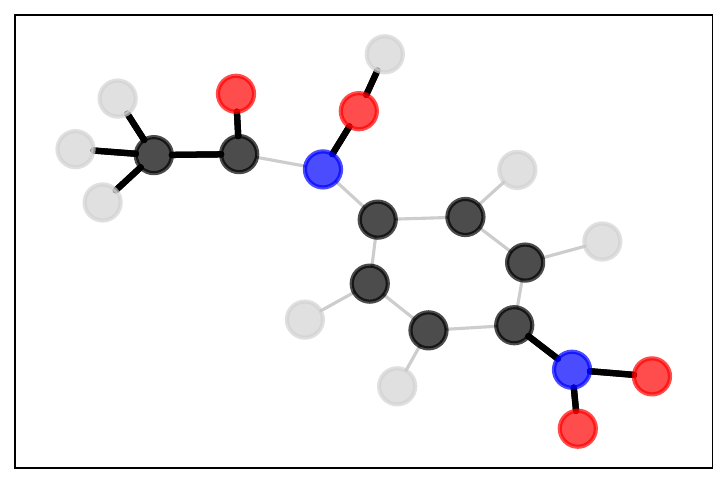}&
\includegraphics[width=5.5em]{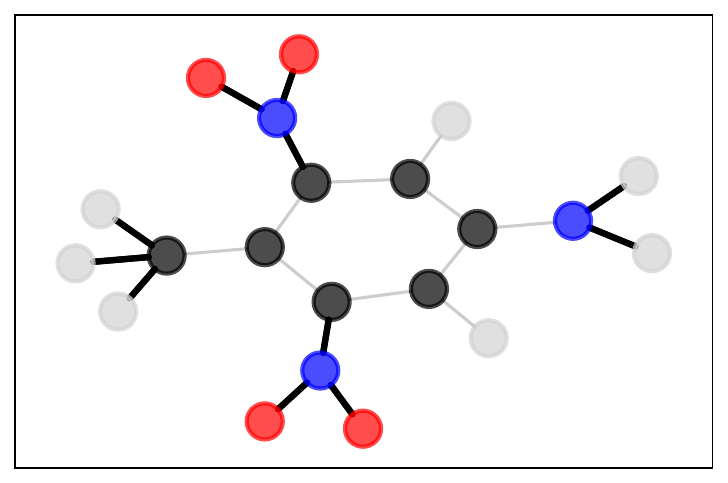}&
\includegraphics[width=5.5em]{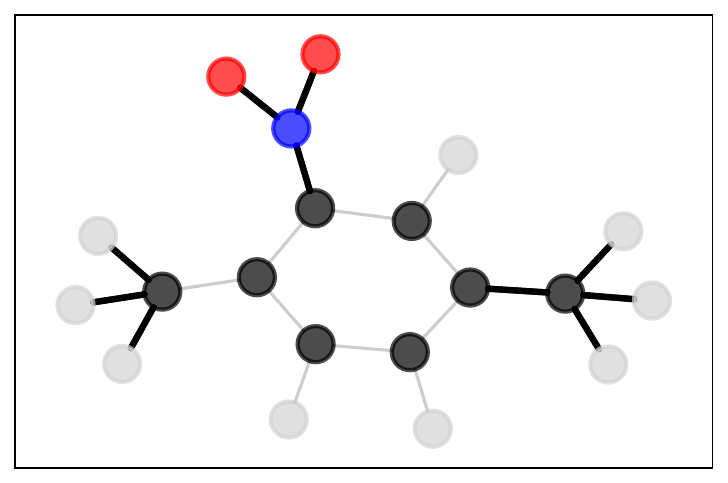} \\ 

\midrule 

GNN-LRP  & 
\includegraphics[width=5.5em]{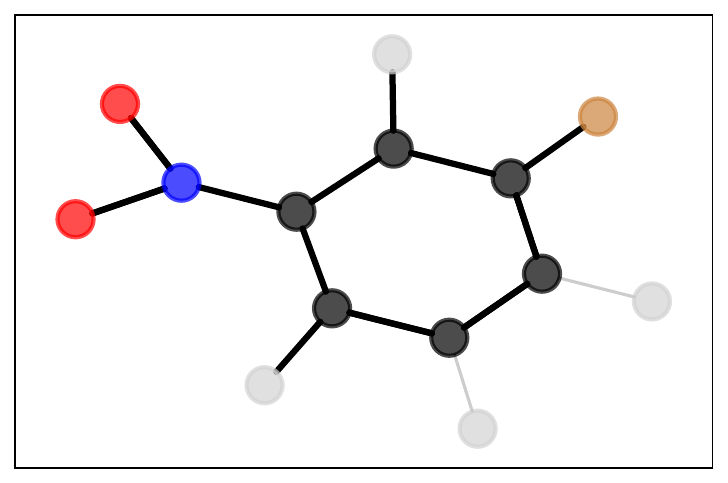}&
\includegraphics[width=5.5em]{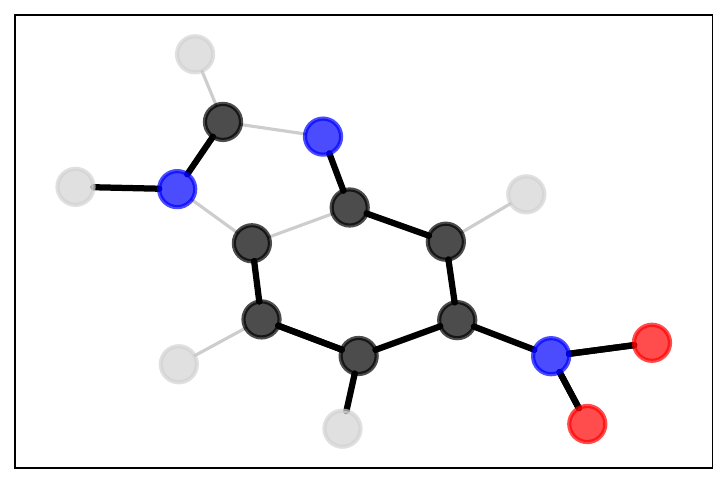}&
\includegraphics[width=5.5em]{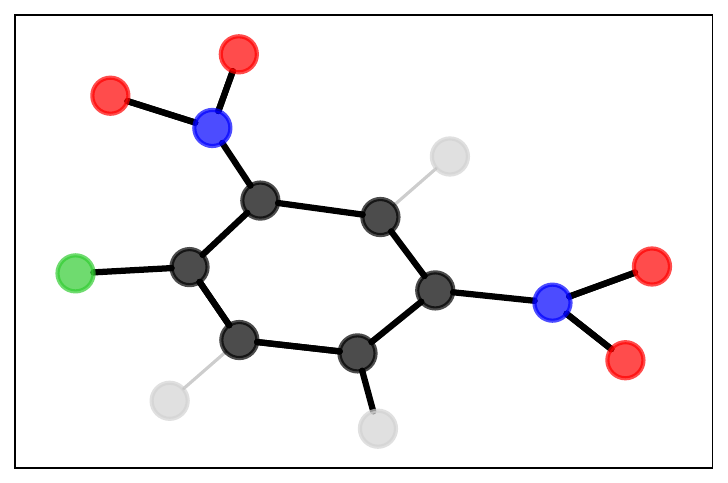}&
\includegraphics[width=5.5em]{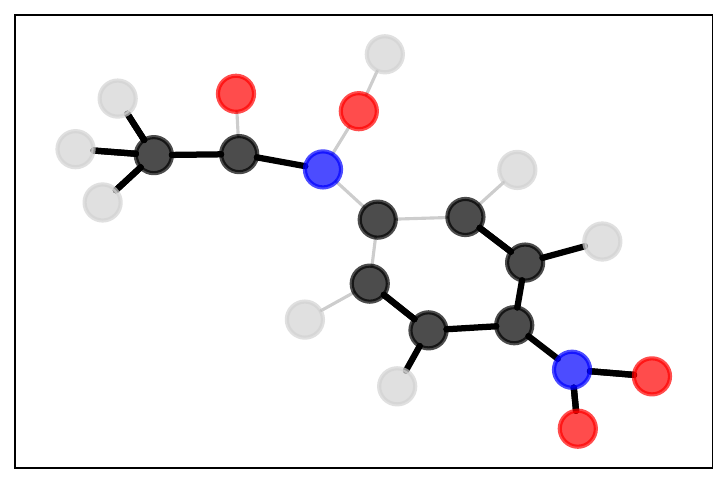}&
\includegraphics[width=5.5em]{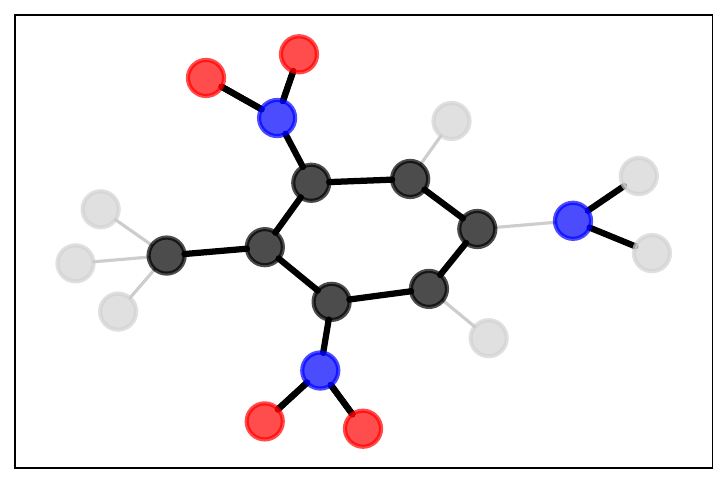}&
\includegraphics[width=5.5em]{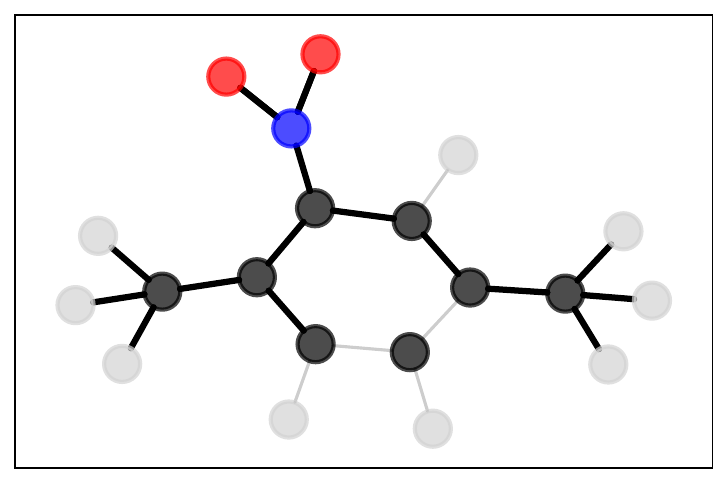} \\ 

\midrule 

\textit{SubgraphX}  & 
\includegraphics[width=5.5em]{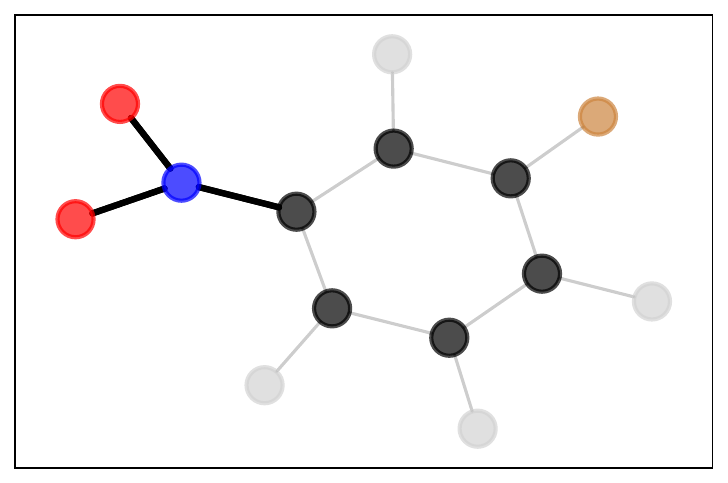}&
\includegraphics[width=5.5em]{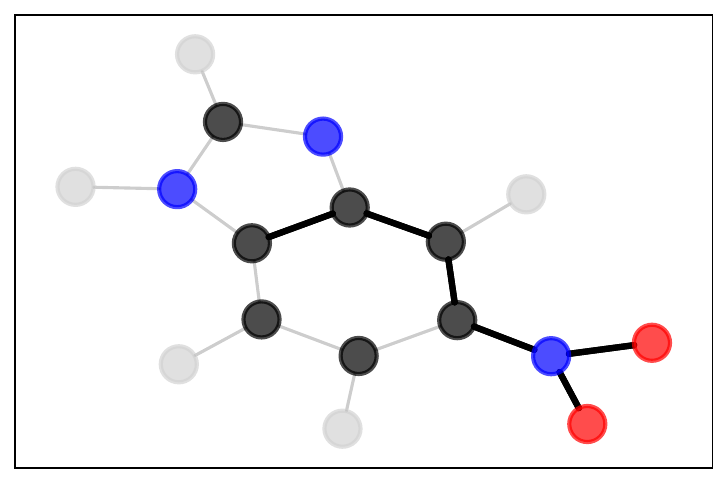}&
\includegraphics[width=5.5em]{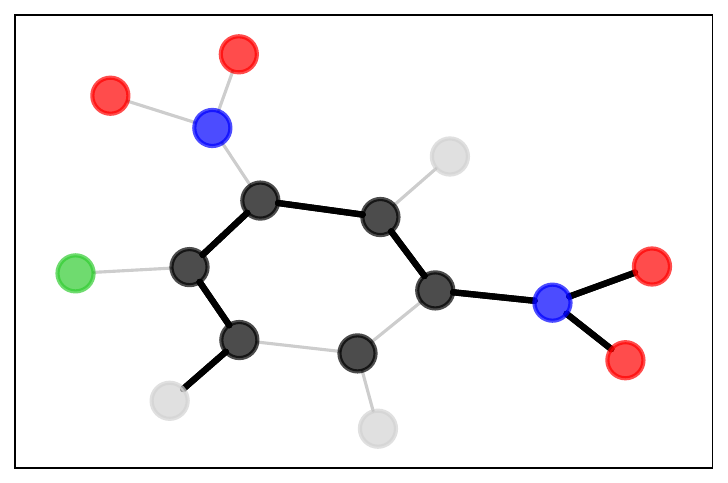}&
\includegraphics[width=5.5em]{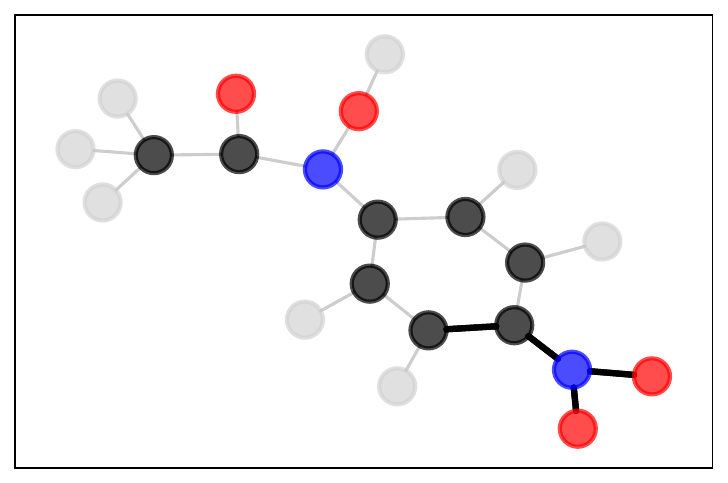}&
\includegraphics[width=5.5em]{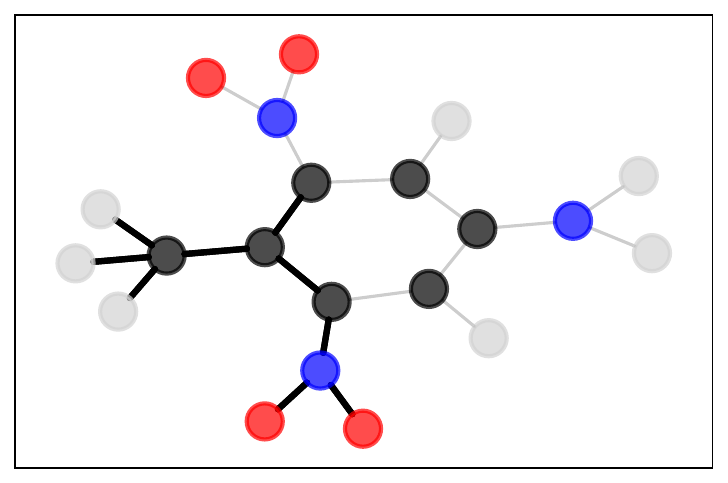}&
\includegraphics[width=5.5em]{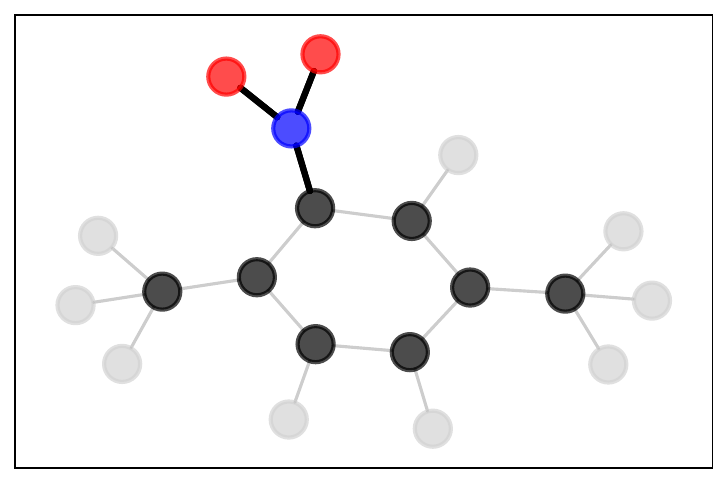} \\ 

\midrule 

GAT & 
\includegraphics[width=5.5em]{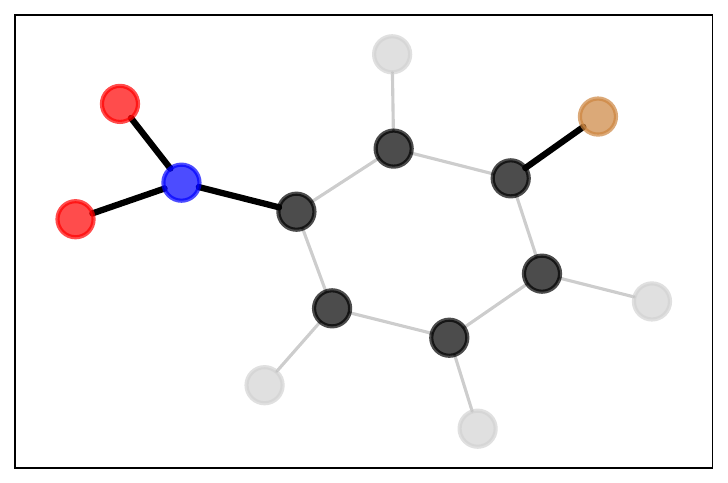}&
\includegraphics[width=5.5em]{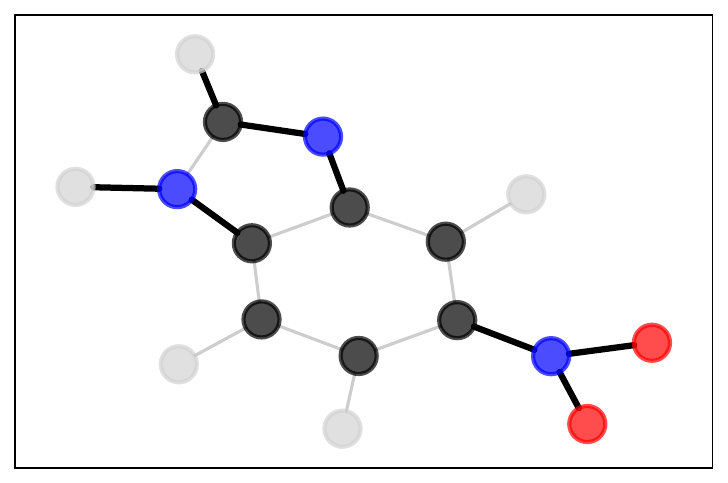}&
\includegraphics[width=5.5em]{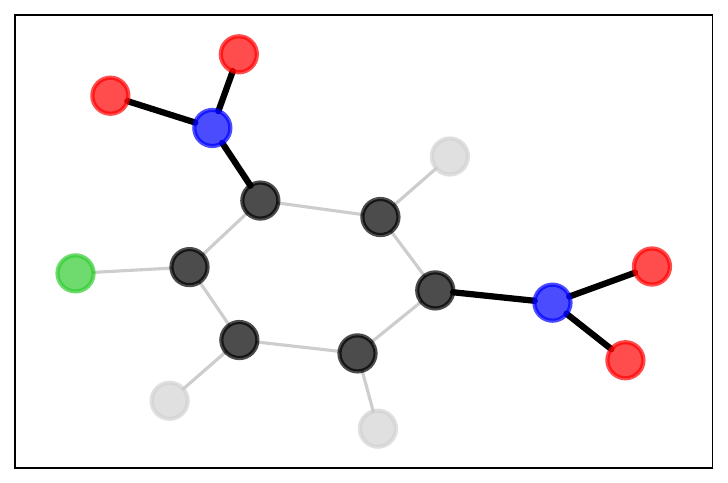}&
\includegraphics[width=5.5em]{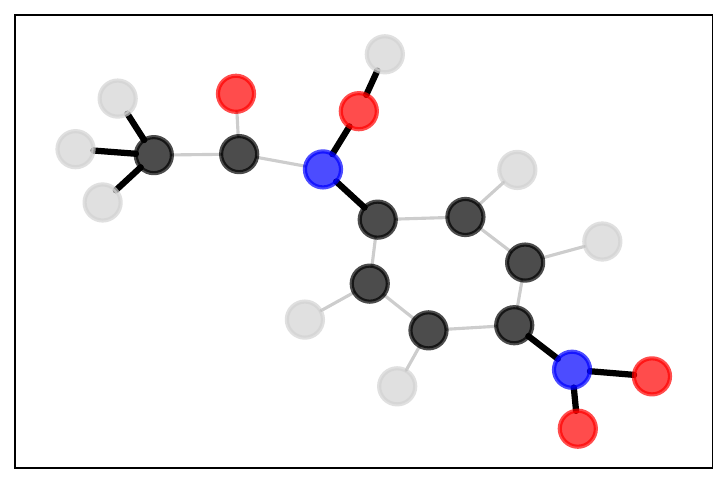}&
\includegraphics[width=5.5em]{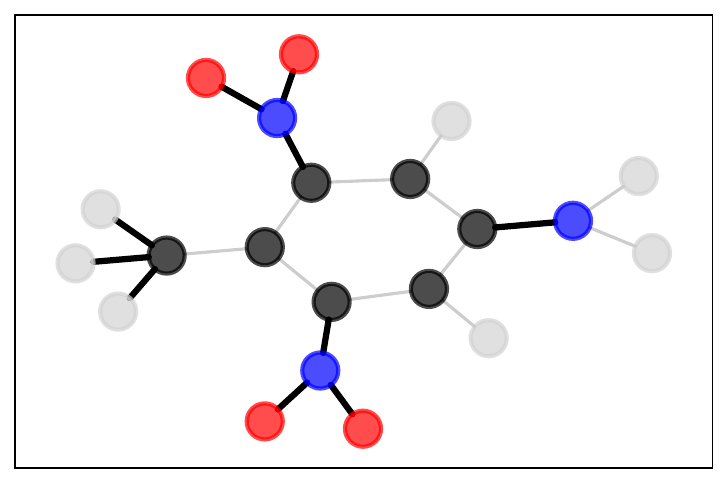}&
\includegraphics[width=5.5em]{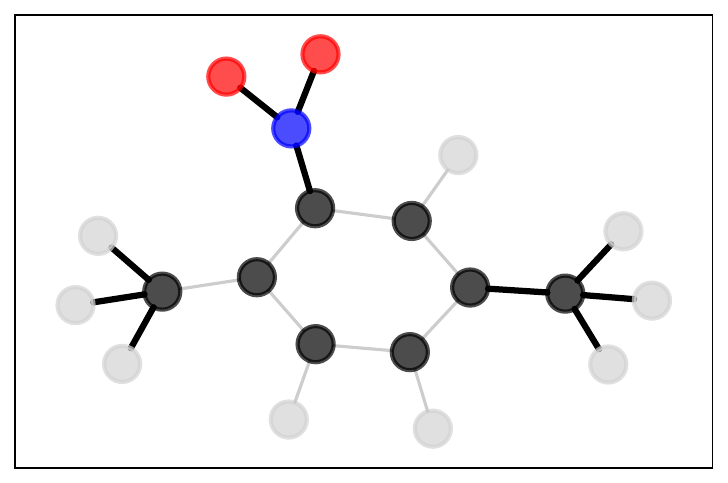} \\ 

\midrule 

$\textsc{L2xGsg}_{dsc}$  & 
\includegraphics[width=5.5em]{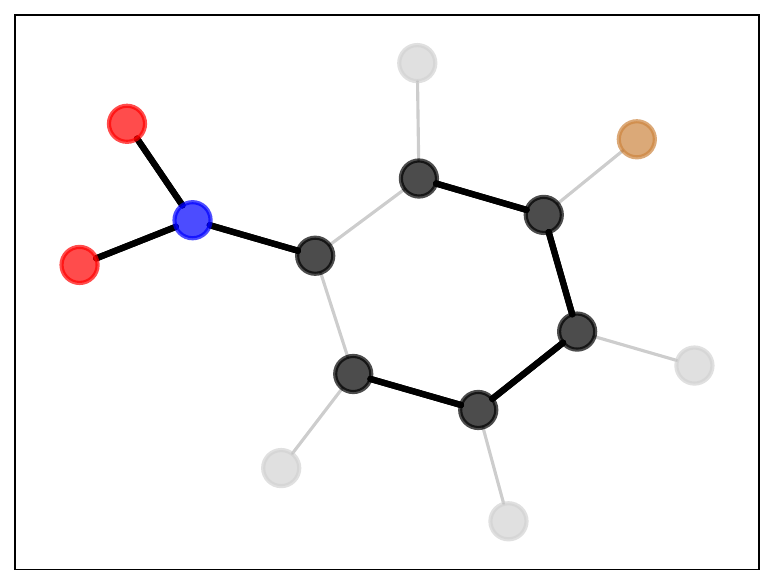}&
\includegraphics[width=5.5em]{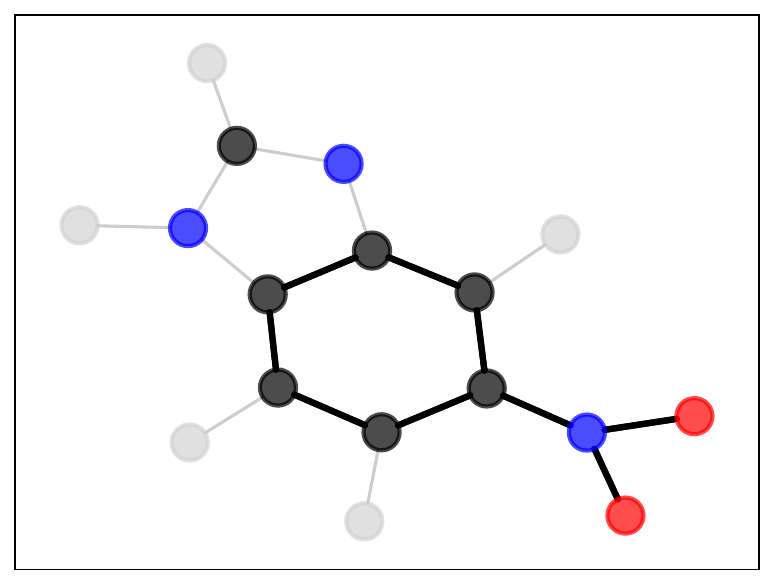}&
\includegraphics[width=5.5em]{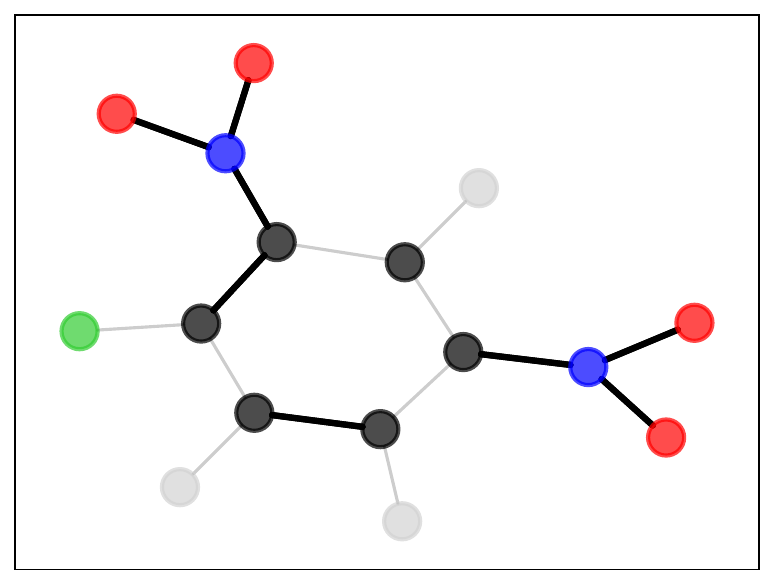}&
\includegraphics[width=5.5em]{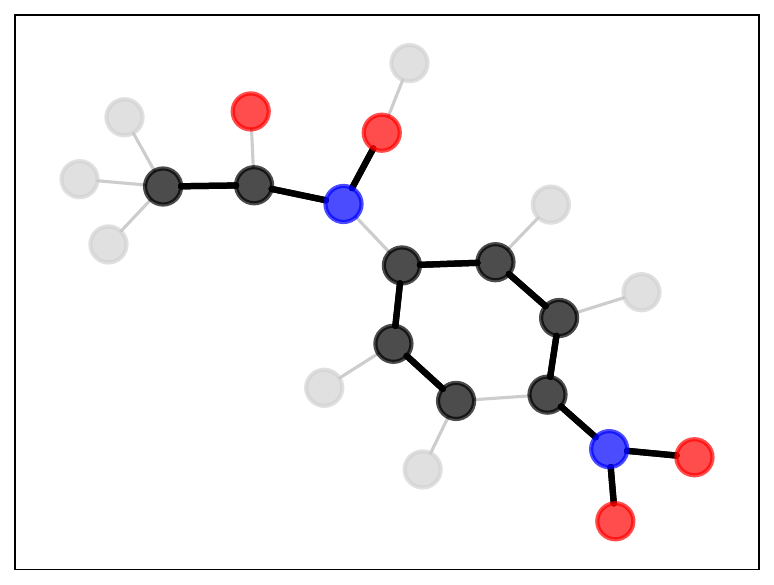}&
\includegraphics[width=5.5em]{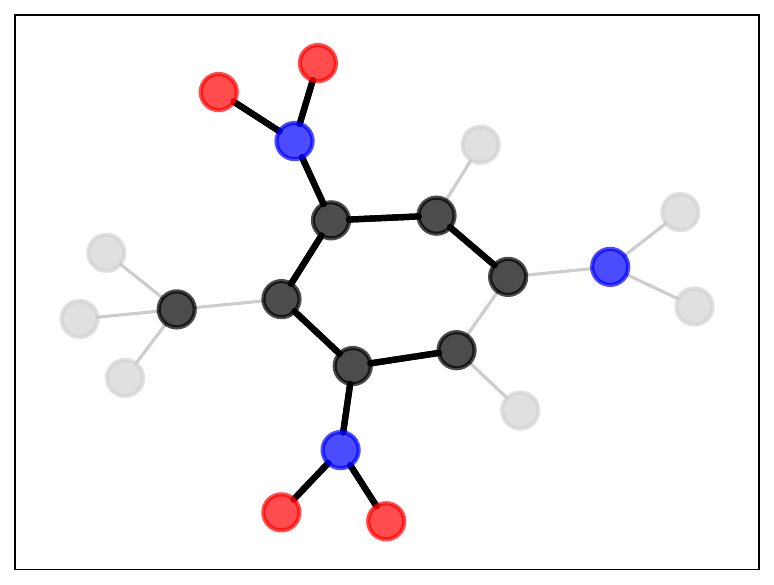}&
\includegraphics[width=5.5em]{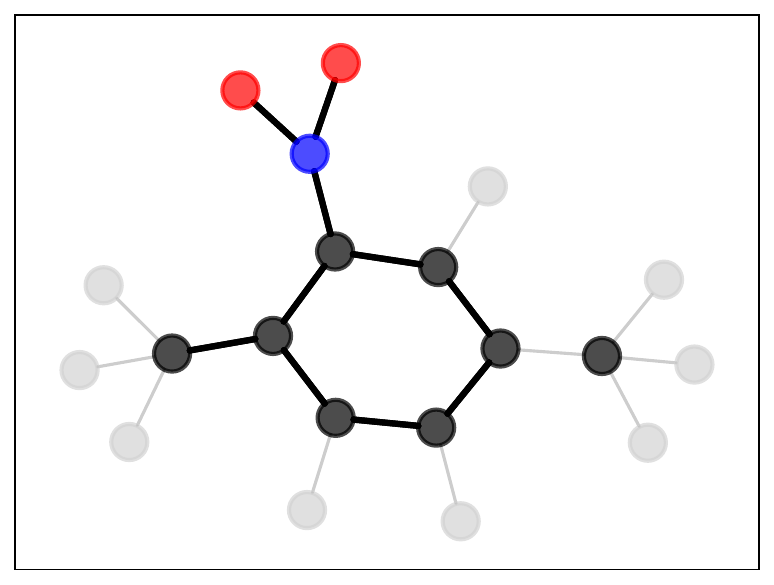} \\ 

\midrule 

\textsc{L2xGsg}  & 
\includegraphics[width=5.5em]{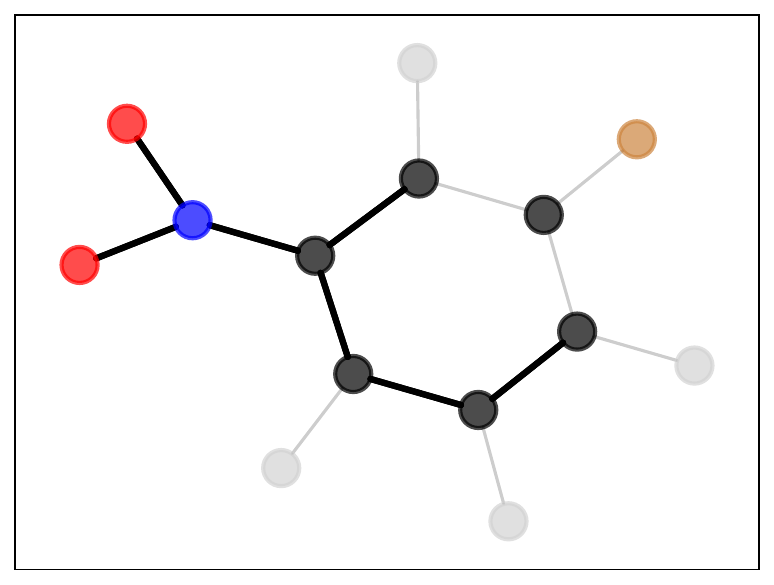}&
\includegraphics[width=5.5em]{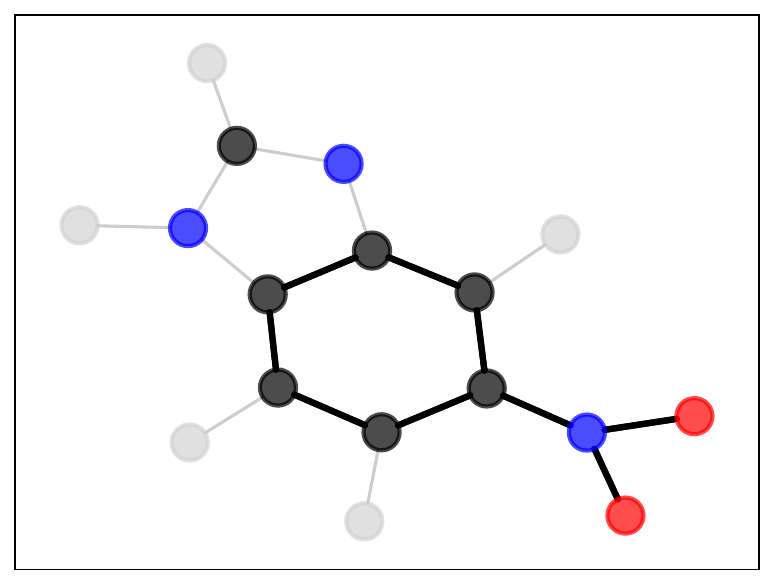}&
\includegraphics[width=5.5em]{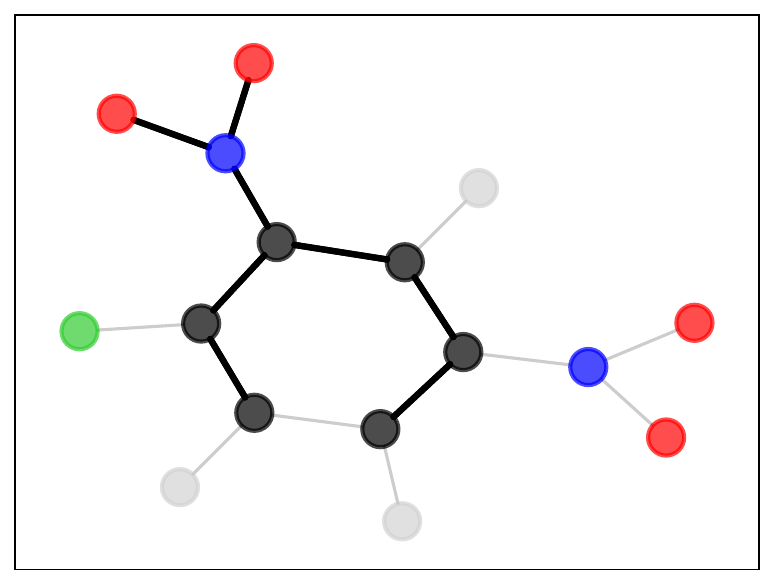}&
\includegraphics[width=5.5em]{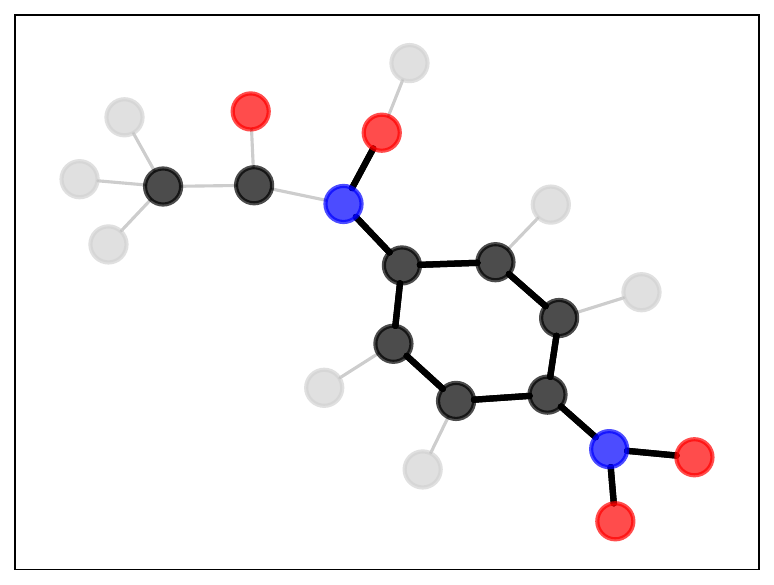}&
\includegraphics[width=5.5em]{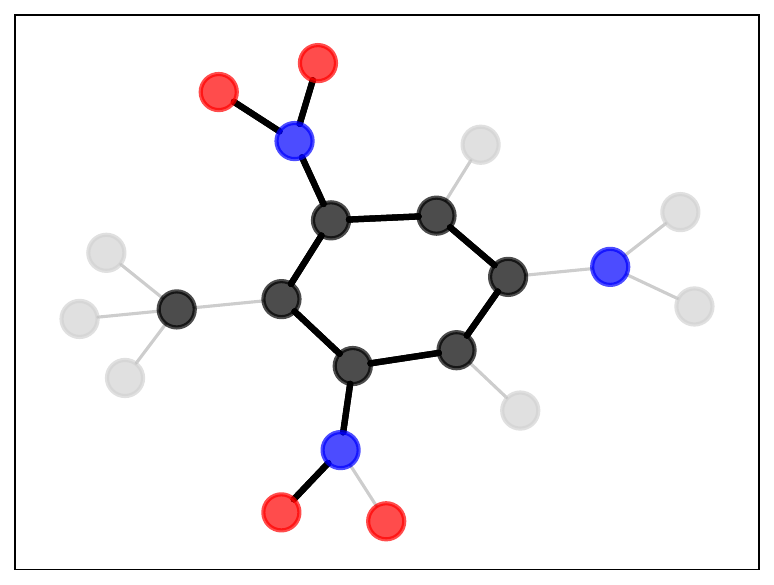}&
\includegraphics[width=5.5em]{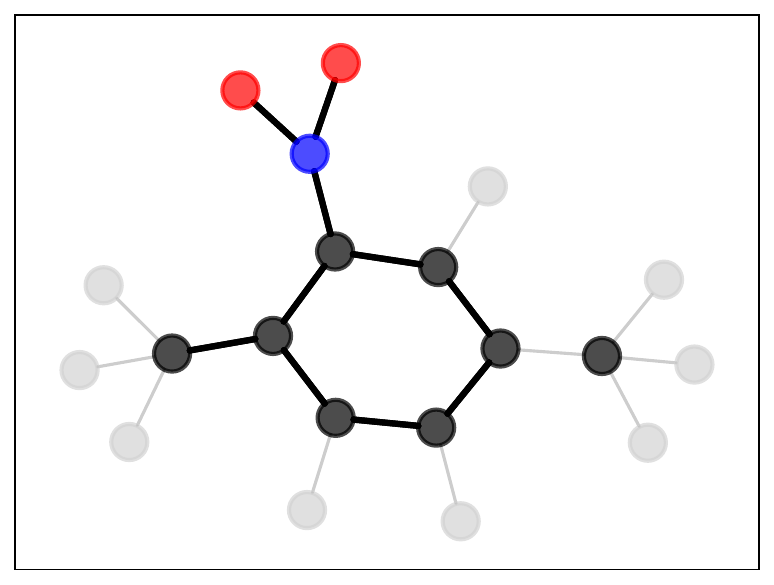} \\ 

\bottomrule 
\end{tabular}
\captionof{figure}{Comparison of the generated explanations for MUTAG$_0$ on the test set. The solid edges are the ones considered responsible of a correct prediction. Black, blue, red, gray, and green nodes represent carbon (C), nitrogen (N), oxygen (O), hydrogen (H), and chlorine (Cl) atoms respectively.}
\label{fig:subgraphs_all}
\end{table} 

\clearpage
\subsection{5x2 Cross-validation Paired t-test} \label{app:5x2}

To prove that our framework can be combined with GNN architectures without hampering their learning capabilities, we perform a paired t-test via 5x2 cross-validation – as suggested in \citet{dietterich1998approximate} – with p-value $0.05$. For completeness, Table \ref{tab:accuracy_5x2test} reports the results of the cross-validation. The paired t-test indicates that there is no statistically significant different in performance between our methods (either connected or not) and the base models. The experiment confirms the findings reported in the main paper and proves that our framework can be integrated into GNN architectures without worrying about performance degradation.

    \begin{table}[h]
        \caption{Prediction test accuracy (\%) for graph classification tasks with 5x2 CV.} \label{tab:accuracy_5x2test}
        
        \begin{tabular}{lccccccccc}
        \toprule
        \multicolumn{1}{l}{\multirow{2}{*}{Method}} &
        \multicolumn{6}{c}{Dataset} \\
        
        \cmidrule(r){2-7}
        \multicolumn{1}{c}{} & \multicolumn{1}{c}{DD} & \multicolumn{1}{c}{MUTAG} & \multicolumn{1}{c}{IMDB-B} & \multicolumn{1}{c}{IMDB-M} & \multicolumn{1}{c}{PROTEINS} & \multicolumn{1}{c}{YEAST} \\

        \cmidrule(r){2-2}\cmidrule(r){3-3}\cmidrule(r){4-4}\cmidrule(r){5-5}\cmidrule(r){6-6}\cmidrule(r){7-7}
        GCN & 
        \textbf{71.4} $\pm$ 1.4 & 
        73.6 $\pm$ 2.9 & 
        73.4 $\pm$ 2.1 & 
        49.2 $\pm$ 1.5 & 
        72.6 $\pm$ 2.1 & 
        87.9 $\pm$ 0.1 \\
        
        $\textsc{L2xGcn}_{dsc}$ & 
        71.2 $\pm$ 1.5 &
        74.2 $\pm$ 3.5 & 
        72.6 $\pm$ 2.2 & 
        \textbf{49.4} $\pm$ 1.2 &
        71.2 $\pm$ 3.4 & 
        \textbf{88.2} $\pm$ 0.2 \\
        
        \textsc{L2xGcn} & 
        71.2 $\pm$ 1.4 & 
        \textbf{75.3} $\pm$ 4.1 &
        \textbf{73.9} $\pm$ 2.1 & 
        49.3 $\pm$ 1.4 & 
        \textbf{73.1} $\pm$ 2.3 & 
        88.0 $\pm$ 0.1 \\
        
        \cmidrule(r){2-2}\cmidrule(r){3-3}\cmidrule(r){4-4}\cmidrule(r){5-5}\cmidrule(r){6-6}\cmidrule(r){7-7}
        GIN & 
        70.7 $\pm$ 1.2 &
        78.2 $\pm$ 8.2 & 
        \textbf{73.1} $\pm$ 2.1 & 
        \textbf{48.9} $\pm$ 1.1 & 
        71.6 $\pm$ 2.4 & 
        \textbf{88.2} $\pm$ 0.1 \\
        
        $\textsc{L2xGin}_{dsc}$ & 
        \textbf{71.6} $\pm$ 1.8 & 
        79.4 $\pm$ 5.7 & 
        71.0 $\pm$ 5.0 & 
        48.5 $\pm$ 1.5 & 
        68.7 $\pm$ 3.3 & 
        88.0 $\pm$ 0.1 \\
        
        \textsc{L2xGin} & 
        71.1 $\pm$ 1.4 & 
        \textbf{79.7} $\pm$ 6.4 & 
        72.3 $\pm$ 2.5 & 
        47.9 $\pm$ 1.4 & 
        \textbf{72.6} $\pm$ 2.5 & 
        88.1 $\pm$ 0.1 \\
        
        \cmidrule(r){2-2}\cmidrule(r){3-3}\cmidrule(r){4-4}\cmidrule(r){5-5}\cmidrule(r){6-6}\cmidrule(r){7-7}
        Graph\textsc{Sage} & 
        71.8 $\pm$ 1.4 & 
        74.4 $\pm$ 4.0 & 
        72.9 $\pm$ 2.2 & 
        \textbf{49.9} $\pm$ 1.2 & 
        71.2 $\pm$ 1.7 & 
        \textbf{88.2} $\pm$ 0.1 \\
        
        $\textsc{L2xGsg}_{dsc}$ & 
        \textbf{71.9} $\pm$ 3.8 & 
        76.1 $\pm$ 2.8 & 
        \textbf{73.1} $\pm$ 1.9 & 
        49.6 $\pm$ 1.1 & 
        \textbf{71.8} $\pm$ 1.5 & 
        88.2 $\pm$ 0.2 \\
        
        \textsc{L2xGsg} & 
        71.8 $\pm$ 1.3 & 
        \textbf{78.2} $\pm$ 5.2 & 
        72.2 $\pm$ 1.8 & 
        49.8 $\pm$ 1.0 & 
        70.1 $\pm$ 3.6 & 
        88.0 $\pm$ 0.2 \\
        \bottomrule
        
        \end{tabular}
        \end{table}




\end{appendices}

\end{document}